\documentclass[10pt,twocolumn,letterpaper]{article}

\usepackage[pagenumbers]{wacv} 

\usepackage{graphicx}
\usepackage{amsmath}
\usepackage{amssymb}
\usepackage{booktabs}

\usepackage{xcolor}
\definecolor{cvprblue}{HTML}{0071BC}
\usepackage[pagebackref,breaklinks,colorlinks,citecolor=cvprblue]{hyperref}

\usepackage[capitalize]{cleveref}
\crefname{section}{Sec.}{Secs.}
\Crefname{section}{Section}{Sections}
\Crefname{table}{Table}{Tables}
\crefname{table}{Tab.}{Tabs.}

\usepackage[labelfont=bf]{caption}
\usepackage{float}
\usepackage{booktabs}
\usepackage{colortbl}
\usepackage{adjustbox}
\usepackage{microtype}
\usepackage{multirow}

\usepackage{mathptmx} 
\usepackage{times}
\usepackage{latexsym}

\usepackage[accsupp]{axessibility} 

\usepackage[percent]{overpic}

\usepackage{pgfplots}
\pgfplotsset{compat=newest}

\usepackage{makecell}
\usepackage{pifont}
\newcommand{\cmark}{\ding{51}}%
\newcommand{\xmark}{\ding{55}}%
\renewcommand{\hline}{\Xhline{3\arrayrulewidth}} 
\newcommand{\thinhline}[1]{\Xhline{\arrayrulewidth}} 

\definecolor{custombboxblue}{HTML}{00A6F6}
\definecolor{customTeal}{HTML}{E90B0C}
\definecolor{customgrey}{HTML}{47494b}
\definecolor{custom_Turquoise}{HTML}{E90B0C}
\definecolor{custom_pink}{HTML}{D95108}
\definecolor{custom_RoyalBlue}{HTML}{1E88E5}
\definecolor{custom_Rust}{HTML}{52352F}
\definecolor{custom_SlateGray}{HTML}{1988C8}
\definecolor{custom_Gold}{HTML}{C48A00}
\definecolor{custom_NavyBlue}{HTML}{303F9F}
\definecolor{custom_Crimson}{HTML}{555555}
\definecolor{custom_Purple}{HTML}{8E24AA}
\definecolor{custom_Lavender}{HTML}{303F9F}
\definecolor{custom_lightBlack}{HTML}{404040}
\definecolor{Green}{RGB}{0, 160, 0} 

\def\wacvPaperID{157} 
\def\confName{WACV}
\def\confYear{2025}

\title{Beyond Boxes: Mask-Guided Spatio-Temporal Feature Aggregation for \\Video Object Detection} 

\makeatletter
\renewcommand{\@fnsymbol}[1]{\ifcase#1\or *\or $\dagger$\else \arabic{#1}\fi}
\makeatother

\author{
Khurram Azeem Hashmi\thanks{Equal contribution.} \quad
Talha Uddin Sheikh\footnotemark[1] \quad
Didier Stricker \quad
Muhammad Zeshan Afzal \\
DFKI - German Research Center for Artificial Intelligence, Kaiserslautern\\
{\tt\small firstname[0]\_firstname[1].lastname@dfki.de}
}

\begin{document}

\twocolumn[{
\maketitle
\begin{center}
    \begin{overpic}[width=\linewidth]{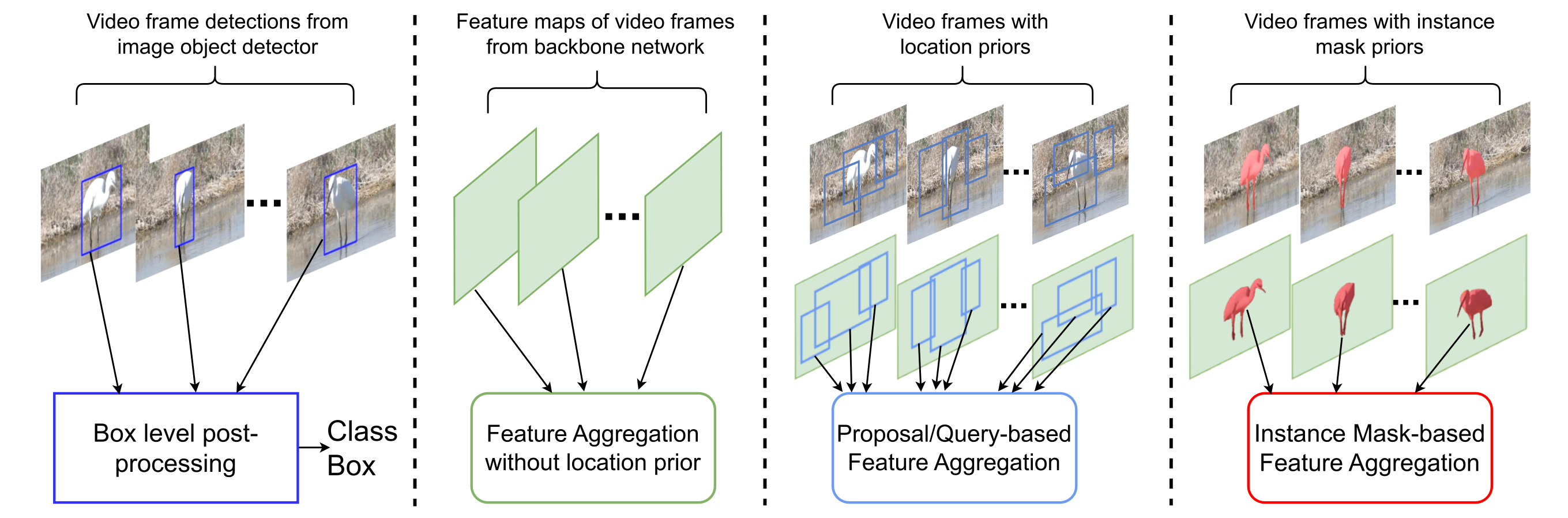}
        \put(3,-2) {\footnotesize\textbf{(a) Box-level}: Seq-NMS~\cite{Seq_NMS_arxiv2016}\label{fig:box_level}}
        \put(28,-2) {\footnotesize\textbf{(b) Frame-level}: FGFA~\cite{Flow_guided_feature_aggregation_ICCV2017}}
        \put(51,-2) {\footnotesize\textbf{(c) Proposal-level}: SELSA~\cite{Sequence_level_semantics_aggregation_ICCV2019}}
        \put(77,-2) {\footnotesize\textbf{(d) Instance Mask-level}: Ours}
      \end{overpic}
      \vspace{0.1pt}
    \captionof{figure}{\textbf{Evolution of exploiting temporal information in video object detection.} \textbf{(a)} Box-level post-processing to refine detections. \textbf{(b)} Feature aggregation across entire video frames. \textbf{(c)} Temporal feature aggregation guided by region-location priors from each frame.  \textbf{(d)} Our instance mask-based aggregation refines the focus to instance boundaries, reducing background noise and improving feature aggregation.}
    \label{fig:evolution_VOD}
\end{center}
}]

\begin{abstract}
\footnotetext[1]{\textsuperscript{*}Equal technical contribution.}The primary challenge in Video Object Detection (VOD) is effectively exploiting temporal information to enhance object representations. Traditional strategies, such as aggregating region proposals, often suffer from feature variance due to the inclusion of background information. We introduce a novel \textbf{instance mask-based feature aggregation} approach, significantly refining this process and deepening the understanding of object dynamics across video frames. We present \textbf{FAIM}, a new VOD method that enhances temporal \textbf{F}eature \textbf{A}ggregation by leveraging \textbf{I}nstance \textbf{M}ask features. In particular, we propose the lightweight Instance Feature Extraction Module (IFEM) to learn instance mask features and the Temporal Instance Classification Aggregation Module (TICAM) to aggregate instance mask and classification features across video frames. Using YOLOX as a base detector, FAIM achieves 87.9\% mAP on the ImageNet VID dataset at 33 FPS on a single 2080Ti GPU, setting a new benchmark for the speed-accuracy trade-off. Additional experiments on multiple datasets validate that our approach is robust, method-agnostic, and effective in multi-object tracking, demonstrating its broader applicability to video understanding tasks. 
\end{abstract}

\section{Introduction}
\label{sec:intro}
Video Object Detection (VOD) aims to identify and locate objects in a video sequence. It has numerous applications, including autonomous driving and video surveillance~\cite{vod_app_autonomous_driving, vod_app_autonomous_driving_2, mot_paper_cvpr_2019}.~Despite remarkable progress in object detection~\cite{Faster_R_CNN_NEURIPS2015,R-FCN_NIPS2016,Fast_RCNN_ICCV2015,You_only_look_once_CVPR2016,Yolov3_arXiv2018,Cascade_RCNN_IEEE2019,Sparse_R-CNN_CVPR2021,Mask_RCNN_ICCV2017, DeformableDETR_arxiv2020, Sparse_R-CNN_CVPR2021, Dynamic_DETR_ICCV2021, Dynamic_sparse_RCNN_CVPR2022, Dino-DETR_Arxiv2022, DETRs_hybrid_matching_CVPR2023}, applying image-based detectors~\cite{Faster_R_CNN_NEURIPS2015, R-FCN_NIPS2016, You_only_look_once_CVPR2016, DeformableDETR_arxiv2020, yolox_arxiv2021} to individual video frames often results in decreased performance. This decline is due to degradation from motion blur, rare poses, camera defocus, and occlusions~\cite{Flow_guided_feature_aggregation_ICCV2017}. However, video frames have the advantage of temporal context, as the detection in one frame can leverage information from surrounding frames. Thus, \textit{effectively exploiting the temporal information in videos} is crucial for addressing the challenges of VOD.

\begin{figure}[t]
    \centering
    \includegraphics[width=\linewidth]{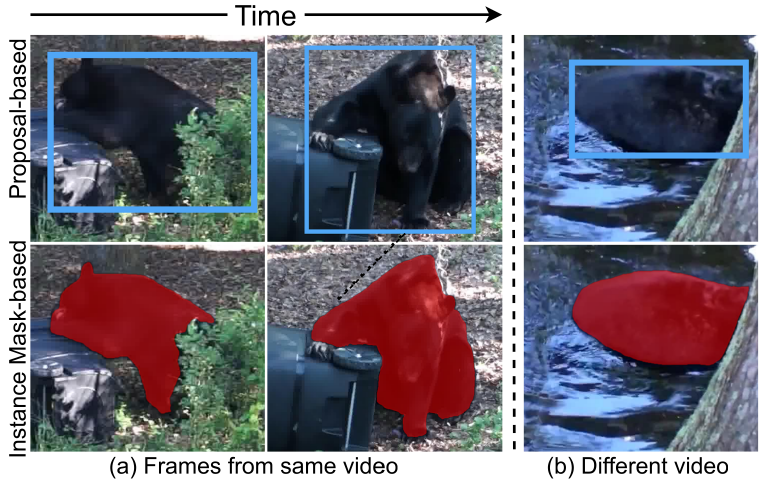}
    \caption{\textbf{Exploiting temporal information} in proposal-based feature aggregation in \textcolor{custombboxblue}{\textbf{blue}} against our instance mask-based feature aggregation method in \textcolor{red}{\textbf{red}} for the class \textbf{Bear}. Leveraging instance mask-level information significantly reduces variance among \textbf{Bear} proposals within and across videos.}
    \label{fig:proposal_mask_diff}
    \vspace{-15pt}
\end{figure}

The exploration of temporal information in VOD has evolved significantly, as illustrated in~\cref{fig:evolution_VOD}. Starting with box-level post-processing~\cite{tcnn_CVPR_2016, T-CNN_IEEE2018,  Detect_to_Track_2017_ICCV, Soft_NMS_ICCV_2017, Seq_NMS_arxiv2016}, progressing through image-level feature aggregation~\cite{Deep_Feature_Flow_CVPR2017, Flow_guided_feature_aggregation_ICCV2017, MANet_Wang_ECCV2018, towards_high_performance_vod_2018_CVPR, TF_blender_ICCV2021, Temporal_meta_adaptor_BMVC2021,spatio_temporal_prompting_2023_ICCV}, and culminating in proposal~(query)-level feature aggregation~\cite{Sequence_level_semantics_aggregation_ICCV2019,RDN_CVPR2019, yao2020video_object_ECCV2020, jiang2020learning_ECCV2020, Memory_enhanced_VOD_CVPR2020, temporal_roi_align_AAAI2021,End-to-end_video_object_detection_ACM2021,sparsevod_bmvc_BMVC2022,muralidhara2022attention,PTSEFormer_ECCV2022, MAMBA_AAAI2021, TransVOD_TPAMI_2022, HVRNet_inter_proposals_ECCV2022, boxmask_2023_WACV,FAQ_2023_CVPR,YOLOV_AAAI2023,objects_disappear_2023_ICCV}. This progression highlights a critical evolution toward enhanced computational efficiency and detection accuracy, focusing on proposal or query-level feature aggregation to reduce background noise and intra-class feature variance compared to image-level aggregation~\cite{Sequence_level_semantics_aggregation_ICCV2019}.
However, this approach remains sub-optimal, as it includes background features, amplifying intra-class variance, particularly in occlusion scenarios (see~\cref{fig:proposal_mask_diff}). This limitation remains a fundamental bottleneck in VOD.

Building on this shift and its limitations, we pose the question: \textit{Can we improve VOD by refining proposal/box-level information to the instance mask-level during temporal feature aggregation}? This paper introduces a novel paradigm in video object detection: instance mask-based feature aggregation. Unlike established proposal-level feature aggregation (Fig.~\ref{fig:evolution_VOD}\textcolor{red}{c}), our approach shown in Fig.~\ref{fig:evolution_VOD}\textcolor{red}{d}, leverages instance pixel-level features to aggregate temporal features across video frames. By focusing on the most granular level—directly around object instances—this method effectively minimizes background noise and intra-class feature variance, as depicted in Fig.~\ref{fig:proposal_mask_diff}. Inspired by recent advancements in box-based semi-supervised instance segmentation~\cite{boxinst_2021_CVPR, BoxTeacher_CVPR2023, lbox2mask_TPAMI_2024}, we introduce the lightweight Instance Feature Extraction Module (IFEM) to learn instance mask features. The Feature Prediction Selection Module (FPSM) refines these features and forwards them to our Temporal Instance and Classification Aggregation Module (TICAM) for final predictions. Additionally, the instance mask features from IFEM are optimized using a mask loss function, comparing them with the pseudo ground truth mask obtained from any box-based instance segmentation methods like Box2Mask~\cite{lbox2mask_TPAMI_2024} or SAM~\cite{SAM_2023}.

\begin{figure}[t]
\begin{tikzpicture}
\begin{axis}[
    width=9cm,  
    height=5cm,  
    title={\textbf{Video Object Detection on ImageNet VID}},
    title style={font=\scriptsize},
    xlabel={Inference Speed},
    ylabel={mAP @50},
    xmin=15, xmax=270,
    ymin=83, ymax=87.5,
    xmode=log,
    log basis x={10},
    ytick={81,82,...,88},
    xtick={15,30, 52, 90, 156, 270},
    xticklabel style={align=center},
    xticklabels={
        15 ms \\(66FPS),
        30 ms \\(33FPS),
        52 ms \\(19FPS),
        90 ms \\(11FPS),
        152 ms \\(6.4FPS),
        270 ms \\(3.7FPS)
    },
    legend pos=north west,
    ymajorgrids=true,
    xmajorgrids=true,
    grid style={line width=0.5pt, draw=gray!30, densely dotted},
    axis line style={draw=none},
    enlarge x limits={rel=0.1},
    enlarge y limits={rel=0.15},
    xlabel style={
        font=\footnotesize 
    },
    ylabel style={
        font=\footnotesize 
    },
    xticklabel style={
        font=\scriptsize 
    },
    yticklabel style={
        font=\scriptsize 
    }
]

    \addplot[only marks, mark=*, mark options={fill=white, draw=customTeal, scale=1.2}] coordinates {
        (90,87.6)
    };
    \addplot[only marks, mark=half-filled-turquoise, mark options={scale=1.2}] coordinates {
        (90,87.4)
    };
    \node[label={[customTeal, align=left, label distance=-0.1cm, font=\fontsize{6.5}{7.2}]right:\textbf{FAIM}}] at (axis cs:90,87.6) {};
    
    \addplot[only marks, mark=triangle*, mark options={fill=white, draw=custom_Crimson, scale=1.2}] coordinates {
        (90,87.1) 
    };
    \node[label={[custom_Crimson, align=left, label distance=-0.1cm, font=\fontsize{6.5}{7.2}]right:\textbf{YOLOV}}] at (axis cs:90,87.1) {};

    \addplot[only marks, mark=pentagon*, mark options={fill=white, draw=custom_SlateGray, scale=1.4}] coordinates {
        ( 90, 86.6) 
    };
    \node[label={[custom_SlateGray, align=left, label distance=-0.1cm, font=\fontsize{6.5}{7.2}]right:\textbf{MAMBA}}] at (axis cs:90,86.6) {};
    
    \addplot[only marks, mark=square, mark options={fill=white, draw=custom_Gold, scale=1.4}] coordinates {
        (90,86.1) 
    };
    \node[label={[custom_Gold, align=left, label distance=-0.1cm, font=\fontsize{6.5}{7.2}]right:\textbf{SELSA}}] at (axis cs:90,86.1) {};
    
    \addplot[only marks, mark=star, mark options={fill=white, draw=custom_pink, scale=1.4}] coordinates {
        (180,87.6) 
    };
    \node[label={[custom_pink, align=left, label distance=-0.1cm, font=\fontsize{6.5}{7.2}]right:\textbf{STPN}}] at (axis cs:180,87.6) {};
    
    \addplot[only marks, mark=x, mark options={fill=white, draw=custom_Purple, scale=1.4}] coordinates {
        ( 180,87.1) 
    };
    \node[label={[custom_Purple, align=left, label distance=-0.1cm, font=\fontsize{6.5}{7.2}]right:\textbf{TROIA}}] at (axis cs:180,87.1) {};
    
    \addplot[only marks, mark=diamond, mark options={fill=white, draw=custom_Lavender, scale=1.4}] coordinates {
        ( 180,86.6) 
    };
    \node[label={[custom_Lavender, align=left, label distance=-0.1cm, font=\fontsize{6.5}{7.2}]right:\textbf{Liu et al.}}] at (axis cs:180,86.6) {};

    \addplot[only marks, mark=+, mark options={fill=white, draw=custom_Rust, scale=1.4}] coordinates {
        (180, 86.1)
    };
    \node[label={[custom_Rust, align=left, label distance=-0.1cm, font=\fontsize{6.5}{8.2}]right:\textbf{TransVOD}}] at (axis cs:180,86.1) {};
    
    \draw (axis cs:81,85.8) rectangle (axis cs:353,87.9);

    \addplot[color=custom_Crimson, line width=1pt, dash pattern=on 2pt off 1pt on 1pt off 1pt] coordinates {
       (16.3,83.6) (22.7,85) (28.9,87.3)
    };
    \addplot[only marks, mark=triangle*, mark options={fill=white, draw=custom_Crimson, scale=1.4}] coordinates {
        (16.3,83.6) (22.7,85) (28.8,87.2)
    };
    
    \addplot[color=customTeal, line width=1pt, dash pattern=on 2pt off 1pt on 1pt off 1pt] coordinates {
       (16.5,84.3) (22.7,85.6)  (28.8,87.9)
    };
    \addplot[only marks, mark=*, mark options={fill=white, draw=customTeal, scale=1.4}] coordinates {
        (16.5,84.3) (22.7,85.6) (28.8,87.9)
    };
    \addplot[only marks, mark=half-filled-turquoise, mark options={draw=customTeal, scale=1.4}] coordinates {
        (11.6,78.2) (16.5,84.3) (22.7,85.6) (28.8,87.9)
    };
    
    \node[label={[custom_Crimson, align=left, label distance=-0.2cm, font=\scriptsize]right:\textbf{L}}] at (axis cs:16.1,82.9) [anchor=south west]{};
    \node[label={[custom_Crimson, align=left, label distance=-0.1cm, font=\scriptsize]right:\textbf{X}}] at (axis cs:22.7,85) {};
    \node[label={[custom_Crimson, align=left, label distance=-0.1cm, font=\scriptsize]right:\textbf{X*}}] at (axis cs:28.8,87.2) {};

    \node[label={[customTeal, align=left, label distance=0.001cm, font=\scriptsize]left:\textbf{L}}] at (axis cs:16.5,84.5) [anchor=south west]{};
    \node[label={[customTeal, align=left, label distance=0.001cm, font=\scriptsize]left:\textbf{X}}] at (axis cs:22.7,85.9) [anchor=south west]{};
    \node[label={[customTeal, align=left, label distance=-0.12cm, font=\scriptsize]left:\textbf{X*}}] at (axis cs:28.8,88.06) {};
    
    \addplot[color=custom_Rust, line width=1pt, dash pattern=on 2pt off 1pt on 1pt off 1pt] coordinates {
        ( 21.9 ,80.3 ) (30.8, 82.3) 
    };
    \addplot[only marks, mark=+, mark options={fill=white, draw=custom_Rust, scale=1.4}] coordinates {
        ( 21.9 ,80.3 ) (30.8, 82.3) 
    };
    \node[label={[custom_Rust, align=center, label distance=0.15cm, font=\scriptsize]right:\textbf{R50}}] at (axis cs: 16.9 ,80.3) [anchor = north west]{};
    \node[label={[custom_Rust, align=center, label distance=0.15cm, font=\scriptsize]right:\textbf{R101}}] at (axis cs: 23.6, 82.3) [anchor = north west]{};
    
    
    \addplot[only marks, mark=+, mark options={fill=white, draw=custom_Rust, scale=1.4}] coordinates {
        (  42.1,83.7) 
    };
    \node[label={[custom_Rust, align=center, label distance=-0.1 cm, font=\scriptsize]below:\textbf{SwinT}}] at (axis cs:  42.1,83.7) {};

    \addplot[color=custom_SlateGray, line width=1pt, dash pattern=on 2pt off 1pt on 1pt off 1pt] coordinates {
         ( 110.3  , 84.6 ) ( 79.6, 83.7)
    };
    \addplot[only marks, mark=pentagon*, mark options={fill=white, draw=custom_SlateGray, scale=1.4}] coordinates {
        ( 110.3  , 84.6 ) ( 79.6, 83.7)
    };
    \node[label={[custom_SlateGray, align=center, label distance=-0.05 cm , font=\scriptsize]left:\textbf{R101}\textsubscript{\textit{ins}}}] at (axis cs:  95.2, 82.7) [anchor=south west]{};
    \node[label={[custom_SlateGray, align=center, label distance=-0.25 cm, font=\scriptsize]\textbf{R101}}] at (axis cs: 110.3, 84.8) [anchor=south]{};

    \addplot[only marks, mark=square, mark options={fill=white, draw=custom_Gold, scale=1.4}] coordinates {
        (  153.8,83.1) 
    };
    \node[label={[custom_Gold, align=center, label distance=-0.1 cm, font=\scriptsize]below:\textbf{X101}}] at (axis cs:  153.8,82.9) {};

    \addplot[only marks, mark=+, mark options={fill=white, draw=custom_NavyBlue, scale=1.4}] coordinates {
        (104.2 ,76.3) 
    };
    \node[label={[custom_NavyBlue, align=left,  label distance=-0.25 cm ,font=\scriptsize]right:\textbf{R101}}] at (axis cs: 104.2 ,76.3  ) [anchor=north west]{};
    
    \addplot[only marks, mark=star, mark options={fill=white, draw=custom_pink, scale=1.4}] coordinates {
        ( 45.7 ,85.0) 
    };
    
    \node[label={[custom_pink, align=center, label distance=-0.0 cm,font=\scriptsize]below:\textbf{SwinT}}] at (axis cs: 40.7 ,85.0) [anchor=south]{};
    
    \addplot[only marks, mark=x, mark options={fill=white, draw=custom_Purple, scale=1.4}] coordinates {
        ( 285.7 ,84.3) 
    };
    \node[label={[custom_Purple, align=center,  label distance=-0.15 cm , font=\scriptsize]below:\textbf{X101}}] at (axis cs: 285.7 ,84.0  ){};
    
    \addplot[only marks, mark=diamond, mark options={fill=white, draw=custom_Lavender, scale=1.4}] coordinates {
        ( 39.6 ,87.2) 
    };
    \node[label={[custom_Lavender, align=center,  label distance=-0.10 cm , font=\scriptsize]right:\textbf{R101}}] at (axis cs: 39.6 ,87.2  ){};
    
    \end{axis}
\end{tikzpicture}
\vspace{-15pt}
\caption{\textbf{Speed and accuracy Trade-off. FAIM outperforms prior state-of-the-art methods on the ImageNet VID benchmark}. Besides QueryProp, MAMBA, and Liu \etal, all results are reported on the 2080Ti GPU. * denotes results with post-processing.}
\label{fig:SOTA_comparison}
\vspace{-15pt}
\end{figure}
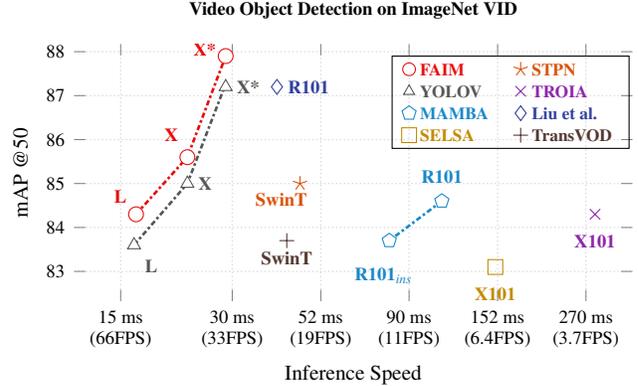

\begin{figure*}[t]
  \centering
  \includegraphics[width=.9\linewidth]{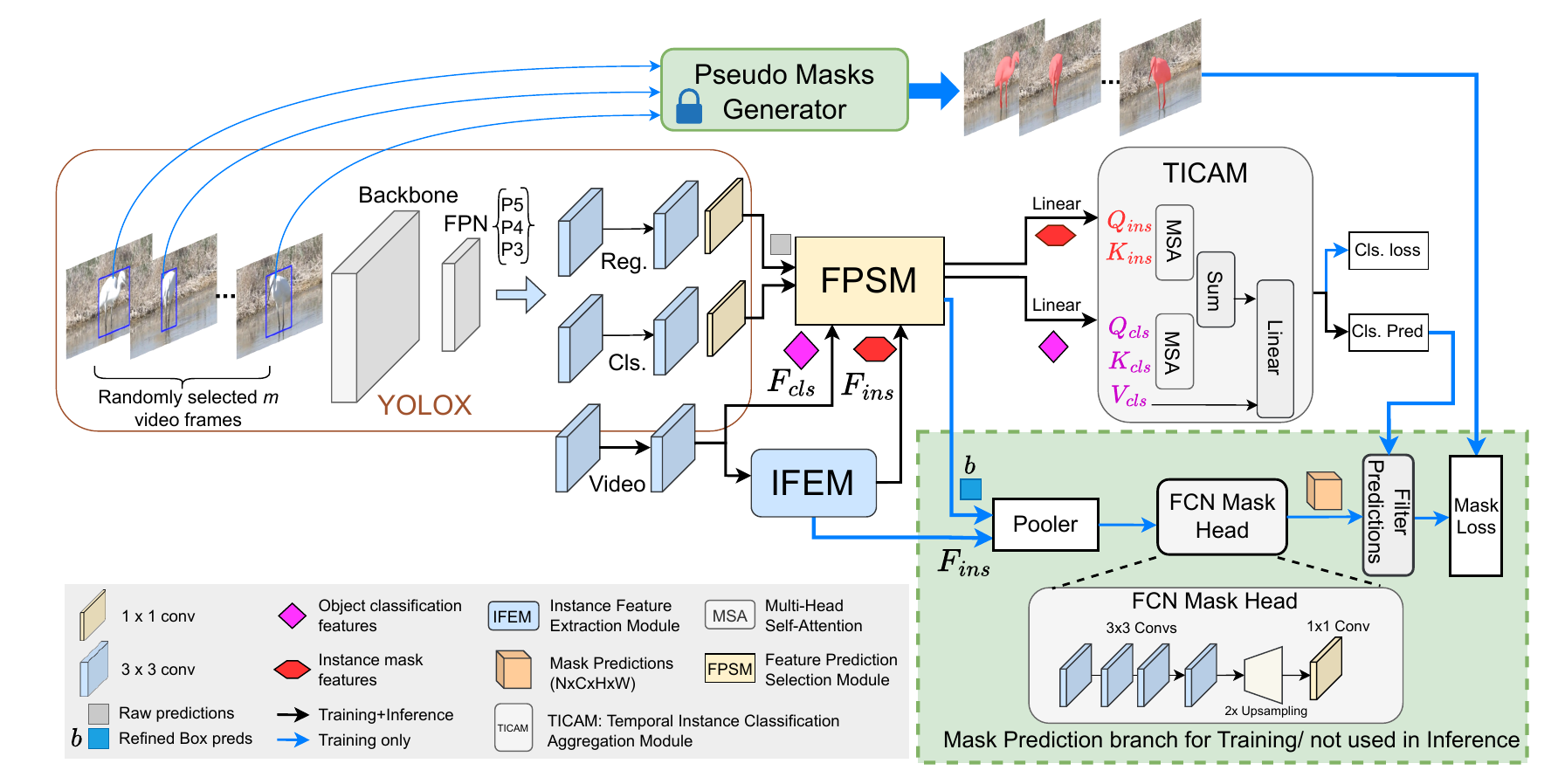}
  \vspace{-10pt}
  \caption{\textbf{Overview of FAIM framework}.Randomly sampled frames from a video are input into YOLOX~\cite{yolox_arxiv2021} for initial feature extraction and prediction using multi-scale features (P3-P5). The IFEM processes video object features to produce instance mask features (Eq.~\ref{eq:ifem}), while the FPSM filters the features for object classification.   IFEM's instance mask features and FPSM's refined predictions are combined to predict instance masks, which are optimized against pseudo-ground truth masks. The learned instance mask features and classification features are then fed into the TICAM for final classification. \textbf{Inference:} Components in \colorbox{green!15}{green} are excluded during inference. However, IFEM continues to provide high-quality instance mask features, enhancing feature aggregation in TICAM for robust predictions.}
  \label{fig:pipeline}
  \vspace{-10pt}
\end{figure*}

Based on these modules, we present \textbf{FAIM}, a new end-to-end VOD framework that enhances temporal \textbf{F}eature \textbf{A}ggregation through leveraging \textbf{I}nstance \textbf{M}ask features.~Following YOLOV~\cite{YOLOV_AAAI2023}, FAIM extends YOLOX~\cite{yolox_arxiv2021} to include the learning of video object and instance mask features with minimal modifications, as shown in Fig.~\ref{fig:pipeline}. Our instance mask-based feature aggregation through FAIM achieves the best speed and accuracy trade-off, as shown in Fig.~\ref{fig:SOTA_comparison}. To summarize, our main contributions are: 

\noindent\textbf{1) Paradigm shift:}~ We introduce a novel paradigm of instance mask-based feature aggregation in VOD, significantly refining the aggregation process and offering a deeper understanding of object dynamics across video frames.

\noindent\textbf{2) FAIM:} The proposed modules in FAIM, such as \textbf{IFEM and TICAM}, are \textbf{method-independent} and can be adapted to other VOD approaches to improve performance (Table~\ref{tab:application_VOD}).
\noindent\textbf{3) Robustness and Generalizability:} Extensive experiments validate that our approach is robust (Tables~\ref{tab:epic_kitchen} and~\ref{tab:ovis}) and applicable to different video understanding tasks, including \textbf{multi-object tracking} (see Table~\ref{tab:application_MOT}).

\section{Related Work}
\label{sec:related_work} 
\noindent\textbf{Box-level Post-Processing.}~Early efforts in video object detection (VOD)~\cite{tcnn_CVPR_2016, T-CNN_IEEE2018, Detect_to_Track_2017_ICCV, Soft_NMS_ICCV_2017, Seq_NMS_arxiv2016} primarily utilized temporal information through box-level post-processing strategies (\cref{fig:evolution_VOD}\textcolor{red}{a}). In these approaches, conventional image object detection methods~\cite{Fast_RCNN_ICCV2015, Faster_R_CNN_NEURIPS2015, R-FCN_NIPS2016, You_only_look_once_CVPR2016} are first applied to individual frames. The predictions from these frames are then refined by integrating temporal cues across sequences. This is achieved through various techniques, including tubelet proposals~\cite{tcnn_CVPR_2016}, tracking~\cite{Detect_to_Track_2017_ICCV}, Soft-NMS~\cite{Soft_NMS_ICCV_2017}, and re-scoring of detections during Non-Maximum Suppression (NMS)~\cite{Seq_NMS_arxiv2016}. While these methods have shown improvements, they do not leverage temporal context during the training phase. Consequently, inaccuracies in initial frame-level detections can propagate throughout the sequence, impacting the overall performance.

\noindent\textbf{Frame-level Feature Aggregation.} Frame-level feature aggregation represents a more sophisticated approach for leveraging temporal information in VOD~\cite{Deep_Feature_Flow_CVPR2017, Flow_guided_feature_aggregation_ICCV2017, MANet_Wang_ECCV2018, towards_high_performance_vod_2018_CVPR, TF_blender_ICCV2021, Temporal_meta_adaptor_BMVC2021} (\cref{fig:evolution_VOD}\textcolor{red}{b}). This methodology begins with feature extraction using backbone networks like ResNet~\cite{Deep_residual_learning_CVPR_2016} and Swin Transformer~\cite{Swin_ICCV_21}, followed by aggregating these features over a temporal window to boost their discriminative capability for the target frame. Pioneering works such as DFF~\cite{Deep_Feature_Flow_CVPR2017} and FGFA~\cite{Flow_guided_feature_aggregation_ICCV2017} employ optical flow fields~\cite{flownet_ICCV2015} to align and aggregate features from adjacent frames. Subsequent advancements have focused on improved feature propagation methods~\cite{MANet_Wang_ECCV2018, towards_high_performance_vod_2018_CVPR} and the effective integration of temporal and spatial features~\cite{TF_blender_ICCV2021, Temporal_meta_adaptor_BMVC2021}. While effective, these methods often overlook long-term temporal dependencies and can be computationally demanding due to frame-level feature aggregation. To overcome these challenges, we propose instance mask-based feature aggregation that not only exploits pixel-level features but also limits the feature aggregation from the image to the instance level.

\noindent\textbf{Proposal/Query-level Feature Aggregation.} Recent advancements in video object detection (VOD) methods~\cite{Sequence_level_semantics_aggregation_ICCV2019, RDN_CVPR2019, yao2020video_object_ECCV2020, jiang2020learning_ECCV2020, Memory_enhanced_VOD_CVPR2020, temporal_roi_align_AAAI2021, End-to-end_video_object_detection_ACM2021, sparsevod_bmvc_BMVC2022, PTSEFormer_ECCV2022, MAMBA_AAAI2021, TransVOD_TPAMI_2022, HVRNet_inter_proposals_ECCV2022, boxmask_2023_WACV} have explored aggregating features at the object proposal level (\cref{fig:evolution_VOD}\textcolor{red}{c}). This approach provides a more context-sensitive and computationally efficient method for incorporating temporal cues. For instance, SELSA~\cite{Sequence_level_semantics_aggregation_ICCV2019}, a pioneering work, aggregates proposal features among video frames based on semantic similarity. TROIA~\cite{temporal_roi_align_AAAI2021} and MEGA~\cite{Memory_enhanced_VOD_CVPR2020} utilize temporal information for extracting and enhancing proposal features, with MEGA introducing a memory-based mechanism to exploit both local and global information. MAMBA~\cite{MAMBA_AAAI2021} introduces a pixel or instance-level memory bank to optimize memory updates for each frame. These methods typically generate proposals on each video frame using a region proposal network~\cite{Faster_R_CNN_NEURIPS2015}. Additionally, query-based feature aggregation methods~\cite{End-to-end_video_object_detection_ACM2021, TransVOD_TPAMI_2022, PTSEFormer_ECCV2022, FAQ_2023_CVPR}, utilizing Transformer-based detectors like Deformable DETR~\cite{DeformableDETR_arxiv2020}, have been explored. Very recently, YOLOV~\cite{YOLOV_AAAI2023} has emerged as a state-of-the-art approach in VOD, balancing speed and accuracy effectively. YOLOV treats detections from a powerful single-stage detector such as YOLOX~\cite{yolox_arxiv2021} as proposals and aggregates features among video frames for final results. These developments highlight the significant gains brought by focusing on objects while aggregating temporal information in VOD. However, all these methods are limited to optimizing box-level information surrounding the object due to the absence of object masks. In contrast, this paper proposes a novel paradigm of instance mask-based feature aggregation, focusing on fine-grained object-level information.

\section{Method}
\label{sec:method}
\noindent\textbf{Overview.} This section first outlines a straightforward approach to transform any proposal-based feature aggregation method to our instance mask-based framework in \S~\ref{subsec: propasl_to_mask_FA}. Building on these principles, we then delve into the design decisions behind FAIM, detailing its unique architectural elements and functionalities in \S~\ref{subsec:FAIM}.

\subsection{From Proposal to Instance Mask-Based Feature Aggregation}
\label{subsec: propasl_to_mask_FA}

\noindent\textbf{Proposal-based Feature Aggregation.}
Let us recall the proposal-based feature aggregation scheme in video object detection~\cite{Sequence_level_semantics_aggregation_ICCV2019,TransVOD_TPAMI_2022, sparsevod_bmvc_BMVC2022, YOLOV_AAAI2023}.~Given an $m$ frames $\{I_1, I_2, \ldots, I_m\}$ from the same video, we first extract feature maps for each frame using a shared backbone network $b_{\text{cnn}}(\cdot;.\theta_{\text{cnn}})$. The feature map $F_t$ for the frame $I_t$ is obtained as:
$F_t$ = $b_{\text{cnn}}(I_t; \theta_{\text{cnn}})$, where $\theta_{\text{cnn}}$ are the parameters of the backbone network.
For each frame $I_t$, a set of proposals $\{P_{t1}, P_{t2}, \ldots, P_{tn_t}\}$ is generated using a Region Proposal Network (RPN)~\cite{Faster_R_CNN_NEURIPS2015} or some image detector~\cite{DeformableDETR_arxiv2020, yolox_arxiv2021}, where $n_t$ is the number of proposals for frame $I_t$. RoIAlign~\cite{Mask_RCNN_ICCV2017} is then applied to extract proposal features $X_{tj}$ for each proposal $P_{tj}$ from the feature map $F_t$:
\vspace{-5pt}
\begin{equation}
X_{tj} = \text{RoIAlign}(F_t, P_{tj}).
\label{eq:roi_feature}
\vspace{-5pt}
\end{equation}
These proposal features $\{X_{t1}, X_{t2}, \ldots, X_{tn_t}\}$ for frame $I_t$ are then aggregated with proposal features from other frames to enhance the feature representation as follows:
\vspace{-5pt}
\begin{equation}
X_{\text{agg}} = \mathcal{A}(\{X_{1j}, X_{2j}, \ldots, X_{mj}\}_{j=1}^{n_t}),
\label{eq:roi_agg_feature}
\vspace{-5pt}
\end{equation}
where $n_t$ is the number of proposals for the frame $I_t$. The aggregation function $\mathcal{A}(\cdot)$ can be a mean, max, or a more complex function like attention-based~\cite{transformer_NIPS2017} feature aggregation.~Before aggregation, these proposal features (used in Eq.~\ref{eq:roi_agg_feature}) are calibrated across space-time based on semantic similarity~\cite{Sequence_level_semantics_aggregation_ICCV2019, temporal_roi_align_AAAI2021}, object classes~\cite{class_aware_IEEECSVT_2022}, memory~\cite{Memory_enhanced_VOD_CVPR2020, MAMBA_AAAI2021}, or prediction confidence~\cite{YOLOV_AAAI2023}. Despite progress, a major limitation is that each proposal feature $X_{tj}$ contains background features in the bounding box. Appearance degradation (common in videos due to rare poses or camera defocus) adversely affects feature aggregation, increasing the intra-class feature variance and decreasing the inter-class feature variance for objects with similar backgrounds. We illustrate this limitation in Fig~\ref{fig:proposal_mask_diff}. To overcome this, we propose using instance mask-based features instead of region proposals during the spatio-temporal feature aggregation. This simple modification isolates object features from the background, reducing intra-class feature variance.

\noindent\textbf{Instance Mask-based Feature Aggregation.}~~Let us consider the same example. After obtaining the RoI features $X_{tj}$ for each proposal $P_{tj}$ using Eq.~\ref{eq:roi_feature}, we propose an Instance Feature Extraction Module (IFEM) to distill instance mask features $M_{tj}$ from proposal features $X_{tj}$:

\begin{equation}
\vspace{-5pt}
M_{tj} = \text{IFEM}(X_{tj})
\label{eq:ifem}
\end{equation}
These instance-mask features $M_{tj}$ are used to predict instance masks (using fully-convolutional head~\cite{R-FCN_NIPS2016, Mask_RCNN_ICCV2017}). Then, these predicted masks are compared against pseudo ground truth masks generated by any box-based instance segmentation methods like Box2Mask~\cite{lbox2mask_TPAMI_2024} or a zero-shot segmentation model like SAM~\cite{SAM_2023}. This comparison refines the instance mask features $M_{tj}$, optimizing them to align closely with the pseudo ground truth masks, thus enhancing the quality of the instance-specific representation.~These instance-mask features can then replace the proposal features in Eq.~\ref{eq:roi_agg_feature} as:

\begin{equation}
\vspace{-5pt}
M_{\text{agg}} = \mathcal{A}(\{M_{1j}, M_{2j}, \ldots, M_{mj}\}_{j=1}^{n_t})
\label{eq:roi_agg_feature_mask}
\end{equation}
This ensures feature aggregation with a higher level of granularity, focusing on the object instances and reducing the background noise.~Thus, \textit{thanks to this simple recipe, any proposal-based feature aggregation scheme can be converted to the instance mask-based feature aggregation approach, without hand-annotated mask labels.} Following this recipe, we propose FAIM to verify the effectiveness of our proposed instance mask-based feature aggregation in VOD.

\subsection{FAIM}
\label{subsec:FAIM}
The FAIM (illustrated in Fig.~\ref{fig:pipeline}) incorporates the instance mask-based feature aggregation in Video Object Detection (VOD). Motivated by the impressive real-time performance of YOLOV~\cite{YOLOV_AAAI2023}, FAIM employs YOLOX~\cite{yolox_arxiv2021} as a base detector with minimal modifications to achieve impressive performance while making it attractive for real-time applications. We now detail each modification.

\noindent\textbf{FPSM: Feature and Prediction Selection Module.}
Initial predictions from the YOLOX~\cite{yolox_arxiv2021} detection head serve as region proposals $\{P_{t1}, P_{t2}, \ldots, P_{tn_t}\}$. Extracting and aggregating features from all these proposals increases computations. Therefore, FPSM filters proposal features and predictions effectively.~Following common conventions~\cite{Faster_R_CNN_NEURIPS2015, Sequence_level_semantics_aggregation_ICCV2019, YOLOV_AAAI2023}, we select top $k$ (e.g., $k$ = 750) predictions based on the confidence scores and perform Non-Maximum Suppression~(NMS) to obtain refined $n$ ($n << k$)  proposals.~Next, to obtain video object-level features, we extend the neck of the base detector with the video object branch as depicted in Fig.~\ref{fig:pipeline}. Similar to the classification and regression branch in~\cite{yolox_arxiv2021}, this branch contains two 3$\times$3 convolutional layers. Unlike YOLOV~\cite{YOLOV_AAAI2023}, which employs the detector's regression features for feature aggregation, our video object branch decouples video object features into classification features $F_{cls}$ and instance mask features $F_{ins}$. The $F_{cls}$ are directly filtered based on the refined predictions in FPSM, whereas $F_{ins}$ are first extracted by our proposed instance feature extraction module. 

\noindent\textbf{IFEM: Instance Feature Extraction Module.}
As depicted in Fig.~\ref{fig:pipeline}, the IFEM is a simple and lightweight module that projects the video object feature into the video instance mask feature space using a single \(3 \times 3\) convolutional layer. Let  \(V^R\in\mathbb{R}^{H \times W \times C}\) represent the video object features extracted from our video object branch, where \(H\), \(W\), and \(C\) denote the height, width, and number of channels, respectively. Using Eq.~\ref{eq:ifem}, IFEM applies a convolution operation \(C(\cdot; \theta_{\text{conv}})\) with parameters \(\theta_{\text{conv}}\) to transform \(V^R\) into instance mask features \(F_{ins} \in \mathbb{R}^{H \times W \times C'}\), with \(C'\) as the number of channels in the transformed feature space. During training, these instance mask features $F_{ins}$ are utilized to predict instance masks, ensuring that the features highly represent the instance masks. Note that the implementation details of IFEM are not important, and even more advanced networks such as~\cite{instance_act_maps_2022_CVPR,mask_frozen_detr_arxiv2023} can be employed. Later, similar to $F_{cls}$, we filter $F_{ins}$ according to refined predictions in FPSM and feed them to the temporal instance classification aggregation module.

\noindent\textbf{TICAM: Temporal Instance Classification Aggregation Module.} Now that we have the filtered video instance mask features and video object classification features, we employ multi-head attention~\cite{transformer_NIPS2017} to aggregate them as explained in Eq.~\ref{eq:roi_agg_feature_mask}.~In our TICAM, the input to multi-head attention includes \(Q_{{cls}}\) and \(Q_{{ins}}\), formed by stacking the features from the classification features $F_{cls}$ and the instance mask features $F_{ins}$ for all proposals across the temporal space (i.e., \(Q_{{cls}}=\text{LP}([F_{{cls}1}, F_{{cls}2}, \ldots, F_{{cls}m}]^T)\) and \(Q_{{ins}}=\text{LP}([F_{{ins}1}, F_{{ins}2}, \ldots, F_{{ins}m}]^T)\). Here, \(\text{LP}(\cdot)\) is the linear projection operator. To verify the effectiveness of the instance mask-based feature aggregation, we adopt the feature aggregation of YOLOV~\cite{YOLOV_AAAI2023} to establish the direct comparison between YOLOV and our FAIM. However, in our TICAM, the temporal aggregation of object classification and instance mask features reduces the background information, producing more discriminative features for VOD. Moreover, it is important to emphasize that our TICAM is independent of the employed feature aggregation scheme. Thanks to Eqs.~\ref{eq:ifem} and~\ref{eq:roi_agg_feature_mask}, it can incorporate other aggregation approaches~\cite{Memory_enhanced_VOD_CVPR2020, MAMBA_AAAI2021, TF_blender_ICCV2021}.

\noindent\textbf{Learning Instance Masks.}
Learning instance masks is a crucial step in our FAIM during training, as shown in Fig.~\ref{fig:pipeline}. In the mask prediction branch, we pool the region features from $F_{ins}$  (from IFEM), according to refined box predictions $b$ (from FPSM), and feed them to Fully Convolutional Network (FCN)~\cite{R-FCN_NIPS2016} to predict instance masks. Here, the Pooler is RoIAlign~\cite{Mask_RCNN_ICCV2017} as explained in Eq.~\ref{eq:roi_feature}. The FCN mask head contains four 3 $\times$ 3 convolutional layers, followed by upsampling and 1$\times$1 convolution to predict mask \(M\in\mathbb{R}^{N \times C \times H \times W}\) with $N$ and $C$ represent the number of proposals and total classes, respectively. $H$ and $W$ denote the size of the predicted mask. For each proposal, we generate \(C\) class-specific predictions. However, comparing \(N \times C\) masks with \(G\) ground truth masks (where \(G \ll N\)) can be sub-optimal during loss computation. Therefore, we use the TICAM's classification outputs to select masks from \(M\) corresponding to positively classified proposals. Formally, let \( P = \{p_1, p_2, \ldots, p_N\} \) be the set of proposals, and the classification predictions from TICAM for these proposals are denoted as \( T = \{t_1, t_2, \ldots, t_N\} \), where \( t_i \in \{1, 2, \ldots, C\} \) represents the predicted class for proposal \( p_i \). The refined mask predictions \( M' \) are obtained by:
\begin{equation}
M' = \{m_i' \mid m_i' = M[i, t_i, :, :], \forall i \in \{1, 2, \ldots, N\},
\label{eq:mask_filter}
\end{equation}
where, \( m_i' \in\mathbb{R}^{H \times W}\) is the mask prediction for proposal \( p_i \) corresponding to its classified category \( t_i \). The proposed filtration approach reduces the number of masks processed and focuses learning on class-specific features, enhancing the network's ability to distinguish between classes. We optimize refined mask predictions $M'$ by minimizing the cross entropy loss, jointly trained in a multi-task fashion~\cite{Mask_RCNN_ICCV2017}, along with detection losses from the base detector~\cite{yolox_arxiv2021}. Again, it is worth mentioning that the implementation details of the mask prediction branch are not important. Here, the goal is not to predict the most accurate segmentation masks but to push $F_{ins}$ to learn instance-specific features. Refer to Appendix~\textcolor{red}{A.2} for the mask loss computation.

\begin{table}[ht]
\vspace{-5pt}
\scriptsize
\centering
\begin{tabular}{@{}l|cccc@{}}
\textbf{Method} & \textbf{Source} & \textbf{Backbone} & \textbf{mAP(\%)$\uparrow$} &\textbf{Time (ms)$\downarrow$ }\\
\hline
SELSA~\cite{Sequence_level_semantics_aggregation_ICCV2019} & ICCV2019& X101 & 83.1 & 153.8 \\
RDN~\cite{RDN_CVPR2019} & ICCV2019& R101 & 81.8 & 162.6 \\
MEGA~\cite{Memory_enhanced_VOD_CVPR2020} &  CVPR2020 & R101 & 82.9 & 230.4 \\
TROIA~\cite{temporal_roi_align_AAAI2021} & AAAI2021 &X101 & 84.3 & 285.7\\ 
MAMBA~\cite{MAMBA_AAAI2021} & AAAI2021 &R101 & 84.6 & 110.3\textit{(T)} \\
QueryProp~\cite{queryProp_AAAI2022} & AAAI2022 &R101 & 82.3 & 30.8\textit{(T)} \\
SparseVOD~\cite{sparsevod_bmvc_BMVC2022} & BMVC2022 &R101 & 81.9 & 142.4 \\
FAQ~\cite{FAQ_2023_CVPR} & CVPR2023 & R50 & 81.7 & 163.2 \\
Liu \etal~\cite{objects_disappear_2023_ICCV} & ICCV2023 &R101 & 87.2 & 39.6\textit{(T)} \\
STPN~\cite{spatio_temporal_prompting_2023_ICCV} & ICCV2023 &SwinT & 85.0 & 45.7 \\
TransVODLite~\cite{TransVOD_TPAMI_2022} & TPAMI2022 & SwinT & 83.7 & 42.1 \\
\hline
YOLOV-S~\cite{YOLOV_AAAI2023} & AAAI2023 & MCSP & 77.3 & 11.3 \\
YOLOV-L~\cite{YOLOV_AAAI2023}  && MCSP & 83.6 & 16.3 \\
YOLOV-X~\cite{YOLOV_AAAI2023}  && MCSP & 85.0 & 22.7 \\
FAIM-S &\cellcolor{custom_RoyalBlue!20}Ours& \cellcolor{custom_RoyalBlue!20}MCSP & \cellcolor{custom_RoyalBlue!20}\textbf{78.2}\textcolor{custom_Turquoise}{\textsubscript{+0.9}} & \cellcolor{custom_RoyalBlue!20}\textbf{11.6} \\
FAIM-L &\cellcolor{custom_RoyalBlue!20}& \cellcolor{custom_RoyalBlue!20}MCSP & \cellcolor{custom_RoyalBlue!20}\textbf{84.3}\textcolor{custom_Turquoise}{\textsubscript{+0.7}} & \cellcolor{custom_RoyalBlue!20}\textbf{16.5} \\
FAIM-X &\cellcolor{custom_RoyalBlue!20}& \cellcolor{custom_RoyalBlue!20}MCSP & \cellcolor{custom_RoyalBlue!20}\textbf{85.6}\textcolor{custom_Turquoise}{\textsubscript{+0.6}} & \cellcolor{custom_RoyalBlue!20}\textbf{22.7}\\
\hline
\textit{With Post-processing}\\
YOLOV-S~\cite{YOLOV_AAAI2023} &AAAI2023& MCSP & 80.1 & \makecell{11.3 $+$ 6.9} \\
YOLOV-L~\cite{YOLOV_AAAI2023} && MCSP & 86.2 & \makecell{16.3 $+$ 6.9} \\
YOLOV-X~\cite{YOLOV_AAAI2023} && MCSP & 87.2 & \makecell{22.7 $+$ 6.1} \\
FAIM-S &\cellcolor{custom_RoyalBlue!20}Ours& \cellcolor{custom_RoyalBlue!20}MCSP & \cellcolor{custom_RoyalBlue!20}\textbf{80.6}\textcolor{custom_Turquoise}{\textsubscript{+0.5}} & \cellcolor{custom_RoyalBlue!20}\textbf{11.6 $+$ 6.9} \\
FAIM-L &\cellcolor{custom_RoyalBlue!20}& \cellcolor{custom_RoyalBlue!20}MCSP & \cellcolor{custom_RoyalBlue!20}\textbf{87.0}\textcolor{custom_Turquoise}{\textsubscript{+0.8}} & \cellcolor{custom_RoyalBlue!20}\textbf{16.5 $+$ 6.9} \\
FAIM-X &\cellcolor{custom_RoyalBlue!20}& \cellcolor{custom_RoyalBlue!20}MCSP & \cellcolor{custom_RoyalBlue!20}\textbf{87.9}\textcolor{custom_Turquoise}{\textsubscript{+0.7}} & \cellcolor{custom_RoyalBlue!20}\textbf{22.7 $+$ 6.9} \\
\bottomrule
\end{tabular}
\caption{\textbf{Comparing accuracy and speed on the ImageNet VID dataset.} \textit{T} denotes the inference time from corresponding papers tested on a different GPU. MCSP is the Modified CSP v5 backbone adopted in YOLOX.  Improvements in \textcolor{custom_Turquoise}{red} highlight gains over YOLOV. \textbf{Our FAIM consistently outperforms YOLOV with all variants of YOLOX while maintaining comparable runtime}.}
\label{tab:performance-comparison}
\vspace{-15pt}
\end{table}


\section{Experiments}
\label{sec:experiments}
\noindent\textbf{Dataset and Evaluation Metrics.} 
Our primary experiments are conducted on the ImageNet VID dataset~\cite{Imagenet_InternationalJournalofComputerVision2015}, comprising 3,862 training videos and 555 validation videos, spanning 30 object classes with annotated bounding boxes.~Adhering to standard VOD protocols~\cite{Sequence_level_semantics_aggregation_ICCV2019, Memory_enhanced_VOD_CVPR2020, TransVOD_TPAMI_2022, YOLOV_AAAI2023}, we utilize a combined dataset of ImageNet VID and DET~\cite{Imagenet_InternationalJournalofComputerVision2015} for training and report results on the validation set using the mean average precision (mAP) metric. Inference runtime is reported in milliseconds (ms) on a single NVIDIA 2080Ti GPU unless stated otherwise.

\noindent\textbf{Base Detector and Backbones.}
Consistent with prior works~\cite{End_to_end_object_detection_spatial_temporal_transformers_ACM2021,TransVOD_TPAMI_2022, yolox_arxiv2021}, we initialize our base detector using COCO pre-trained weights from YOLOX~\cite{yolox_arxiv2021}.~Our FAIM is evaluated across different YOLOX variants (YOLOX-S, YOLOX-L, YOLOX-X), each incorporating the Modified CSP v5 backbone~\cite{CSPNet_cvprw2020}. Consequently, we refer to our FAIM variants as FAIM-S, FAIM-L, and FAIM-X.

\noindent\textbf{Training.}~To directly compare with YOLOV~\cite{YOLOV_AAAI2023}, we adopt an identical training strategy and follow the original codebase\footnote{https://github.com/YuHengsss/YOLOV} from the authors. We sample one-tenth of the frames from the ImageNet VID training set to address the redundancy. The base detectors are trained as in~\cite{YOLOV_AAAI2023} with a batch size of 16 on 2 GPUs. When base detectors are integrated into our FAIM, we fine-tune them on a batch size of 16 on a single GPU. The same learning schedule is adopted, and only the newly added video object feature branch, instance feature extraction module, FCN mask head, and multi-head attention are fine-tuned. To generate pseudo ground truth instance masks, we try pre-trained SAM with the ViT-H~\cite{VIT_ICLR2021} image encoder and pre-trained Box2Mask~\cite{lbox2mask_TPAMI_2024} with ResNet-101~\cite{Deep_residual_learning_CVPR_2016} backbone network. Owing to better performance, we select ground truth instance masks from SAM for experiments. Refer to Appendix~\textcolor{red}{C.1} for the performance comparison between SAM and Box2Mask. During training, in the FPSM, the NMS is set to 0.75 to select box predictions and features. In TICAM, the number of frames $m$ is set to 16. 

\noindent\textbf{Testing.} During testing, the NMS threshold is set to 0.5, whereas the number of frames $m$ for feature aggregation is empirically set to 32. Complete implementation details are provided in Appendix~\textcolor{red}{A.1}. 

\begin{figure*}[ht]
  \centering
  \includegraphics[width=.95\linewidth]{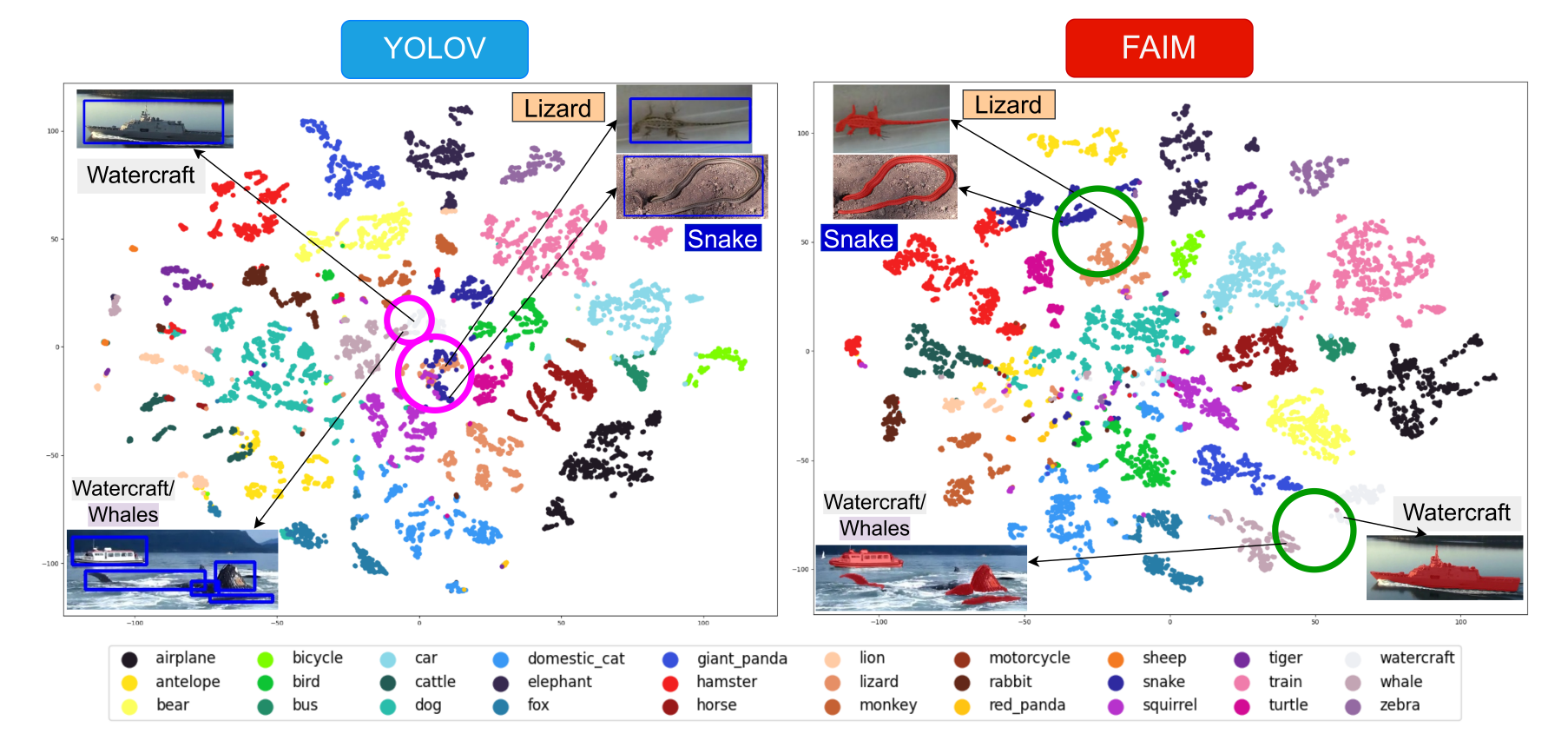}
  \vspace{-15pt}
  \caption{TSNE of proposal features from YOLOV~\cite{YOLOV_AAAI2023} and FAIM on the ImageNet VID dataset. Feature confusion in YOLOV is marked with \textcolor{magenta}{magenta circles} $\color{magenta}\bigcirc$, and corrections in FAIM with \textcolor{Green}{green circles} $\color{Green}\bigcirc$. The \textcolor{blue}{blue bounding box} \textcolor{blue}{\fbox{}} shows the area used for feature aggregation in YOLOV, while FAIM uses the area in \textcolor{red}{red mask}. YOLOV confuses features between \textit{\textbf{Snake}} and \textit{\textbf{Lizard}} (highlighted with $\color{magenta}\bigcirc$), showing higher intra-class and lower inter-class variance due to background inclusion. FAIM's instance mask-based feature aggregation reduces this variance, forming clearer clusters. Similar improvements are seen with \textbf{\textit{Watercraft}} and \textbf{\textit{Whale}}. Best viewed on a screen.}   \label{fig:tsne_proposals}
  \vspace{-5pt}
\end{figure*}

\begin{table*}[t]
\begin{minipage}[t]{0.29\linewidth}
\centering
\footnotesize
\begin{adjustbox}{width=\linewidth}
\begin{tabular}{@{}l|ccccc@{}}
\hline
\vphantom{\text{\large$m_{g}$} $\rightarrow$}Method & {IFEM} & {TICAM} &  {T (ms)} & {mAP} \\
YOLOV-S & \xmark & \xmark   &11.30 & 77.3  \\
\rowcolor{custom_RoyalBlue!20}YOLOV-S & \cmark & \xmark &11.60 & 77.9\textsubscript{\textcolor{custom_Turquoise}{\textbf{+0.6}}} \\
YOLOX-S & \xmark & \xmark   &9.40 & 69.5 \\
\rowcolor{custom_RoyalBlue!20}FAIM-S & \cmark & \cmark & 11.60  & \textbf{78.2}\textsubscript{\textcolor{custom_Turquoise}{\textbf{+8.9}}} \\
\end{tabular}
\end{adjustbox}
\vspace{-5pt}
\caption{\textbf{Effectiveness of the modules proposed in FAIM}.}
\label{tab:FAIM_module_ablations}
\end{minipage}
\hfill
\begin{minipage}[t]{0.27\linewidth}
\centering
\footnotesize
\begin{adjustbox}{width=\linewidth}
\begin{tabular}{@{}l|cccccc@{}}
\hline
\text{\large$m_{g}$} $\rightarrow$ & 3 & 7 & 15 & 23 &  \cellcolor{custom_RoyalBlue!20}{31} & 39\\
\hline
mAP & 75.4 & 76.8 & 77.7 & {77.9} &  \cellcolor{custom_RoyalBlue!20}{\textbf{78.2}} & 78.2\\
\hline
\text{\large$m_{l}$} $\rightarrow$ & 3 & 7 & 15 & 23 & 31 & 39\\
\hline
mAP & 71.8 & 72.6 & 73.4 & 73.8 & 74.3 & 74.6\\
\end{tabular}
\end{adjustbox}
\vspace{-5pt}
\caption{\textbf{Varying global $m_{g}$ and local reference frames $m_{l}$}.}
\label{tab:ablation_frame_sampling}

\end{minipage}
\hfill
\begin{minipage}[t]{0.33\linewidth}
\centering
\begin{adjustbox}{width=\linewidth}
\begin{tabular}{@{}l|cccccc@{}}
\hline
\vphantom{\text{\large$m_{g}$} $\rightarrow$}\text{\large$n$} $\rightarrow$ & 10 & 20 &  \cellcolor{custom_RoyalBlue!20}30 & 50 & 75 & 100\\
\hline
mAP & 76.9 & 77.8 &  \cellcolor{custom_RoyalBlue!20}{\textbf{78.2}} &  78.3 & 78.4 & 78.4\\
Time (ms)  & 10.68 & 10.98 & \cellcolor{custom_RoyalBlue!20}{\textbf{11.60}} & 14.17 & 20.02 & 30.08\\
\end{tabular}
\end{adjustbox}
\vspace{-5pt}
\caption{\textbf{Investigating different number of proposals $n$ in FPSM.}}
\label{tab:proposals_number}
\end{minipage}
\vspace{-15pt}
\end{table*}


\subsection{Main Results}
\label{sec:main_results}
Our FAIM aims for real-time video object detection (VOD). Therefore, we mainly compare it with several state-of-the-art methods focussing on real-time VOD. As shown in Table~\ref{tab:performance-comparison}, we present a quantitative analysis comparing both mAP (\%) and the inference run-time of FAIM against other prominent VOD methods~\cite{Sequence_level_semantics_aggregation_ICCV2019,RDN_CVPR2019, Memory_enhanced_VOD_CVPR2020,temporal_roi_align_AAAI2021,MAMBA_AAAI2021,queryProp_AAAI2022,FAQ_2023_CVPR,sparsevod_bmvc_BMVC2022,TransVOD_TPAMI_2022,YOLOV_AAAI2023, spatio_temporal_prompting_2023_ICCV,objects_disappear_2023_ICCV}. YOLOV, our direct competitor with the same detector and backbone, is compared in all three variants. Thanks to our novel instance mask-based feature aggregation and the efficiency of the single-stage detector, FAIM consistently and significantly surpasses the previous state-of-the-art, specifically YOLOV~\cite{YOLOV_AAAI2023}, achieving the highest mAP of 87.9\% and 85.6\% with and without sequential post-processing~\cite{Robust_efficient_post_processing_Video_Object_Detection_IROS2020}, respectively. Notably, our lightweight instance feature extraction module results in a negligible increase in inference run-time (+0.3 ms and +0.2 ms compared to YOLOV-S and YOLOV-L, respectively). However, it brings considerable gains of +0.9\% and +0.7\% in mAP without post-processing. When adopting a larger detector like YOLOX-X, the difference in run-time becomes negligible, while the mAP improvement remains significant at +0.7\%.~Apart from~\cite{MAMBA_AAAI2021,queryProp_AAAI2022,objects_disappear_2023_ICCV}, all models are evaluated on the same GPU for a direct comparison. Moreover, it is worth noting that the proposed modules in FAIM are method-agnostic and can be plugged into other VOD methods to improve performance (see Table~\ref{tab:application_VOD}).


\noindent\textbf{Qualitative Comparison.} We extract and compare the proposal features from the FAIM's Temporal Instance Classification Aggregation Module (TICAM) and YOLOV's Feature Aggregation Module (FAM)\cite{YOLOV_AAAI2023} using t-SNE in Fig.\ref{fig:tsne_proposals}. As demonstrated, FAIM's use of instance mask-level features offers a significant advantage, as it leads to compact clustering of proposals within each class, reducing intra-class variance. Moreover, it increases the separation between different classes, particularly among visually similar or background-heavy categories, such as \textit{Watercraft} and \textit{Whale}. This improvement allows FAIM to better differentiate between objects with overlapping contexts or similar backgrounds. Appendix~\textcolor{red}{B} offers more qualitative analysis.

\begin{table*}[t]

\begin{subtable}{0.25\linewidth}
\centering
\begin{adjustbox}{width=\linewidth}
\begin{tabular}{@{}l|cccc@{}}
Scale $\rightarrow$ & P3 & P4 & \cellcolor{custom_RoyalBlue!20} P5 & P3-P5\\
\hline
mAP & 78.1 & 77.9 &\cellcolor{custom_RoyalBlue!20}{\textbf{78.2}}& 78.2\\
\end{tabular}
\end{adjustbox}
\caption{\textbf{ \# FPN scale for RoIAlign.}}
\label{tab:sclae_roialign}
\vspace{-7pt}
\end{subtable}
\hfill
\begin{subtable}{0.25\linewidth}
\centering
\begin{adjustbox}{width=\linewidth}
\begin{tabular}{@{}l|cccc@{}}
Size $\rightarrow$ & 14$\times$ 14 & 28 $\times$ 28  &  \cellcolor{custom_RoyalBlue!20}32$\times$32\\
\hline
mAP & 77.7& 78.1 &\cellcolor{custom_RoyalBlue!20}{\textbf{78.2}}\\
\end{tabular}
\end{adjustbox}
\caption{\textbf{RoIAlign output Size.}}
\label{tab:roi_size}
\vspace{-7pt}
\end{subtable}
\hfill
\begin{subtable}{0.30\linewidth}
\centering
\begin{adjustbox}{width=\linewidth}
\begin{tabular}{@{}l|cc@{}}
Mask loss$\rightarrow$ & \cellcolor{custom_RoyalBlue!20} Class Aware & Class Agnostic\\
\hline
mAP  &\cellcolor{custom_RoyalBlue!20}{\textbf{78.2}}& 77.8\\
\end{tabular}
\end{adjustbox}
\caption{\textbf{Instance Mask loss computation.}}
\label{tab:class_aware}
\vspace{-7pt}
\end{subtable}
\hfill
\begin{subtable}{0.15\linewidth}
\centering
\begin{adjustbox}{width=\linewidth}
\begin{tabular}{@{}l|cc@{}}
Loss $\rightarrow$ & Dice &  \cellcolor{custom_RoyalBlue!20}BCE\\
\hline
mAP & 77.9  & \cellcolor{custom_RoyalBlue!20}{\textbf{78.2}}\\
\end{tabular}
\end{adjustbox}
\caption{\textbf{Loss Function.}}
\label{tab:mask_loss_function}
\vspace{-7pt}
\end{subtable}
\caption{\textbf{Ablating mask prediction branch in FAIM}. Settings for results in \S~\ref{sec:experiments} are \colorbox{custom_RoyalBlue!20}{highlighted}.}
\label{tab:mask_branch_ablation}
\vspace{-10pt}
\end{table*}


\begin{table}[ht]
\centering
\scriptsize
\begin{tabular}{@{}l|lll@{}}
Method& {mAP\textsubscript{50}} & {mAP\textsubscript{75}} & {mAP\textsubscript{50:95}} \\
\hline
SELSA*~\cite{Sequence_level_semantics_aggregation_ICCV2019}  & 78.4 & 52.5 & 48.6 \\
SELSA+Ours  & \cellcolor{custom_RoyalBlue!20}\textbf{79.5\textcolor{custom_Turquoise}{\textsubscript{+1.1}}} & \cellcolor{custom_RoyalBlue!20}\textbf{54.4\textcolor{custom_Turquoise}{\textsubscript{+1.9}}} & \cellcolor{custom_RoyalBlue!20}\textbf{49.6\textcolor{custom_Turquoise}{\textsubscript{+1.0}}} \\
\hline
TROIA*~\cite{temporal_roi_align_AAAI2021}  & 78.9 & 52.8 & 48.8 \\
TROIA+Ours& \cellcolor{custom_RoyalBlue!20}\textbf{80.1\textcolor{custom_Turquoise}{\textsubscript{+1.2}}} & \cellcolor{custom_RoyalBlue!20}\textbf{55.4\textcolor{custom_Turquoise}{\textsubscript{+2.6}}} & \cellcolor{custom_RoyalBlue!20}\textbf{50.0\textcolor{custom_Turquoise}{\textsubscript{+1.2}}} \\
\end{tabular}
\vspace{-5pt}
\caption{\textbf{Exploring instance mask-based feature aggregation in other VOD methods.} Results with * are reproduced. Improvement of over 1\% is observed.}
\label{tab:application_VOD}
\end{table}

\begin{table}[ht]
    \centering
    \scriptsize
\begin{tabular}{l|l|l}
    Method & AP50/AP75 (S1) & AP50/AP75 (S2)\\
    \hline 
    Liu~\cite{objects_disappear_2023_ICCV}& 44.9/18.7 & 41.7/16.0 \\
    TROIA~\cite{temporal_roi_align_AAAI2021}& 42.2/13.3 & 39.6/11.3 \\
    \rowcolor{custom_RoyalBlue!20}TROIA+Ours & \textbf{45.1/18.9}~\textcolor{custom_Turquoise}{+\textbf{2.9}/\textbf{+5.6}} & \textbf{42.0/16.2}~\textcolor{custom_Turquoise}{+\textbf{2.4}/\textbf{+4.9}}
\end{tabular}
    \vspace{-5 pt}
    \caption{Results on EPIC KITCHENS-55~\cite{Epic_kitchens_ECCV2018}. S1 and S2 are seen and unseen splits. We achieve new SOTA results.}
    \label{tab:epic_kitchen}
\end{table}

\begin{table}[ht]
    \centering
    \scriptsize
    \begin{tabular}{l|l|l|l}
         Method & AP & AP50 & AP75 \\
         \hline
         YOLOV-X~\cite{YOLOV_AAAI2023} & 54.7 & 75.0 & 57.2\\
         \rowcolor{custom_RoyalBlue!20}FAIM-X & \textbf{55.8}~\textcolor{custom_Turquoise}{\textbf{+1.1}} & \textbf{76.9}~\textcolor{custom_Turquoise}{\textbf{+1.9}} & \textbf{58.6}~\textcolor{custom_Turquoise}{\textbf{+1.4}}
    \end{tabular}
    \vspace{-5 pt}
    \caption{Our FAIM achieves \textbf{stronger gains of +1.9 points in AP50} on the OVIS~\cite{OVIS_IJCV_2022} dataset with severe occlusions.}
    \label{tab:ovis}
    \vspace{-10 pt}
\end{table}

\subsection{Ablation Studies}
\label{sec:ablations}
We analyze the design decisions in FAIM using YOLOX-S as a base detector on the validation set of ImageNet VID~\cite{Imagenet_InternationalJournalofComputerVision2015} dataset. We employ similar settings to Sec.~\ref{sec:experiments} and report performance on the standard mAP\(_{50}\) and runtime in milliseconds (ms). More ablations studies and evaluations are provided in Appendix~\textcolor{red}{C}.

\noindent\textbf{Effectiveness of each component.} Table~\ref{tab:FAIM_module_ablations} analyzes the contributions of our proposed modules IFEM and TICAM on both YOLOV-S~\cite{YOLOV_AAAI2023} and YOLOX-S~\cite{yolox_arxiv2021}. For YOLOV-S, the baseline achieves a mAP of 77.3\%. Adding IFEM results in a mAP increase of +0.6\%, bringing the total to 77.9\%, with a negligible runtime increase (+0.3ms). This improvement suggests that even without incorporating instance mask features into the feature aggregation module, the addition of instance mask learning in~\cite{YOLOV_AAAI2023} helps refine the temporal object classification queries $Q_{cls}$ for better classification. In YOLOX-S, originally a single-frame detector, adding both IFEM and TICAM transforms it into FAIM-S, a video object detection model. This yields a substantial improvement, increasing mAP from 69.5\% to 78.2\% (+8.9\%) with a modest runtime increase of +2.2ms. IFEM introduces instance mask learning, while TICAM effectively aggregates temporal mask and classification features across frames, reducing feature variance and significantly improving detection performance. A comparison of TICAM and IFEM with standard attention-based methods~\cite{Dynamic_head_CVPR2021} can be found in Appendix~\textcolor{red}{C.4}.

\noindent\textbf{Reference frame sampling.}~
Consistent with previous research~\cite{Sequence_level_semantics_aggregation_ICCV2019, temporal_roi_align_AAAI2021, boxmask_2023_WACV, YOLOV_AAAI2023},
we explore both global and local frame sampling strategies in our work. The results in Table~\ref{tab:ablation_frame_sampling} reveal that using merely 3 global reference frames surpasses the performance achieved with 39 local reference frames. This finding is in line with prior works~\cite{Sequence_level_semantics_aggregation_ICCV2019, temporal_roi_align_AAAI2021, boxmask_2023_WACV, YOLOV_AAAI2023}. Therefore, in alignment with the approach in~\cite{YOLOV_AAAI2023}, we adopt the global sampling strategy with $m_{l}$=31 as the default.

\noindent\textbf{Number of Proposals.} We study the effect of varying the number of prediction proposals $n$ from 10 to 100 in FPSM. As shown in Table~\ref{tab:proposals_number}, our approach, FAIM, demonstrates a notable increase of 0.9\% in mAP when \( n \) increases from 10 to 20. This performance already surpasses that of YOLOV-S~\cite{YOLOV_AAAI2023} (with \( n = 30 \)) by \textbf{+0.5\%} in mAP, while also being faster by \textbf{0.6 milliseconds}. Further elevating \( n \) to 30 results in an additional mAP gain of +0.4\%, albeit with an increase of 0.6 milliseconds in runtime. The improvement continues consistently as \( n \) is increased, reaching a plateau at \( n = 75 \). Given the quadratic complexity (\( O(n^2) \)) of the self-attention mechanism in TICAM, we opt for \( n = 30 \).\\
\noindent\textbf{Design Choices for Mask Prediction.} Table~\ref{tab:mask_branch_ablation} presents an ablation study of the mask prediction branch in FAIM. We analyze the impact of pooling features from different scales (P3-P5) in the model's neck (see Fig.~\ref{fig:pipeline}), as detailed in Eq.~\ref{eq:roi_feature}. Table~\ref{tab:sclae_roialign} shows that pooling features from P5 yields the best mask prediction results. Hence, P5 is used by default. Table~\ref{tab:roi_size} explores varying the RoIAlign output size, with 32$\times$32 chosen for optimal performance during training, as mask prediction is not required during inference. Table~\ref{tab:class_aware} demonstrates that filtering mask predictions based on TICAM's classification improves mAP by 0.4\%. Table~\ref{tab:mask_loss_function} shows that Binary Cross-Entropy (BCE) loss is the most effective for mask loss and is used by default. The mask prediction branch in FAIM is modular and can be fully modified. Further ablations are presented in Appendix~\textcolor{red}{C}.
\subsection{FAIM in other two-stage VOD Methods}
\label{sec:app_vod}
\noindent\textbf{Settings.}
Following the recipe detailed in \S~\ref{subsec: propasl_to_mask_FA}, this study evaluates the adaptability of our instance mask-based feature aggregation in two-stage, proposal-based VOD methodologies, namely SELSA~\cite{Sequence_level_semantics_aggregation_ICCV2019} and TROIA~\cite{temporal_roi_align_AAAI2021}. Following the 1x schedule in MMTracking~\cite{mmtrack2020} with ResNet-50 as the backbone, we examine these methods with and without our instance mask-based feature aggregation scheme (see Eq.~\ref{eq:ifem}). Implementation details are available in Appendix~\textcolor{red}{A.2}.

\noindent\textbf{Results.} 
Table~\ref{tab:application_VOD} lists the results, demonstrating that the integration of instance mask-based feature aggregation yields a significant \textbf{improvement of more than 1\% in mAP\textsubscript{50}} for both SELSA~\cite{Sequence_level_semantics_aggregation_ICCV2019} and TROIA~\cite{temporal_roi_align_AAAI2021}. Notably, these enhancements are achieved with minimal modifications, as detailed in \S~\ref{sec:experiments}. These outcomes in Table~\ref{tab:application_VOD} confirm the efficacy of the instance mask-based feature aggregation technique in two-stage proposal-based VOD methods, suggesting its potential for further improvements.

\subsection{Additional VOD Benchmarks}
\label{sec:add_exp}

\noindent\textbf{Experiments on EPIC KITCHENS-55.} Besides ImageNet VID, we report results on the more challenging EPIC KITCHENS-55 dataset~\cite{Epic_kitchens_ECCV2018}, comprising ego-centric videos of 32 different kitchens and 290 classes. Implementation details are in Appendix~\textcolor{red}{A.3}. Table~\ref{tab:epic_kitchen} summarizes the results. When our proposed instance mask-based feature aggregation is integrated into TROIA~\cite{temporal_roi_align_AAAI2021}, we surpass prior state-of-the-art results in both splits, affirming its applicability to challenging video object detection tasks.

\noindent\textbf{Experiments on OVIS.} Following the experimental setting in~\cite{YOLOV_AAAI2023}, we compare the performance of our FAIM and YOLOV on the Occluded Video Instance Segmentation (OVIS) dataset~\cite{OVIS_IJCV_2022}. This dataset contains 25 classes and is notable for its high level of occlusion, with many objects being partially or fully occluded in multiple frames. Refer to Appendix~\textcolor{red}{A.4} for more implementation details. As shown in Table~\ref{tab:ovis}, FAIM-X surpasses YOLOV-X by a significant margin, highlighting the effectiveness and robustness of our instance mask-based feature aggregation on occluded VOD tasks.

\begin{table}[t]
\centering
\scriptsize
\begin{tabular}{@{}l|cccc@{}}
Method & MOTA~\cite{clear_MOT_metrics_2008}$\uparrow$ & {IDF1}~\cite{IDF1_ECCV2016}$\uparrow$ & HOTA~\cite{HOTA_IJCV2021}$\uparrow$ &{IDS}~\cite{clear_MOT_metrics_2008}$\downarrow$ \\
\hline
Tracktor*~\cite{Traktor_2019_ICCV}  & 70.5 & 65.3 & 53.0 & 1442 \\
Tracktor+Ours  & 
\cellcolor{custom_RoyalBlue!20}\textbf{71.4\textcolor{custom_Turquoise}{\textsubscript{+0.9}}} & \cellcolor{custom_RoyalBlue!20}\textbf{66.7\textcolor{custom_Turquoise}{\textsubscript{+1.4}}} & \cellcolor{custom_RoyalBlue!20}\textbf{53.1\textcolor{custom_Turquoise}{\textsubscript{+0.1}}} & \cellcolor{custom_RoyalBlue!20}\textbf{1344\textcolor{custom_Turquoise}{\textsubscript{-98}}} \\
\hline
ByteTrack*~\cite{bytetrack_ECCV2022} & 86.4 & 82.7 & 65.5 & 995 \\
ByteTrack+Ours& 
\cellcolor{custom_RoyalBlue!20}\textbf{88.1\textcolor{custom_Turquoise}{\textsubscript{+1.7}}} & \cellcolor{custom_RoyalBlue!20}\textbf{83.7\textcolor{custom_Turquoise}{\textsubscript{+1.0}}} & \cellcolor{custom_RoyalBlue!20}\textbf{68.9\textcolor{custom_Turquoise}{\textsubscript{+3.4}}} & \cellcolor{custom_RoyalBlue!20}\textbf{911\textcolor{custom_Turquoise}{\textsubscript{-84}}} \\
\end{tabular}
\vspace{-5pt}
\caption{\textbf{Exploring instance mask-based learning in Multi-Object Tracking.} Results with * are reproduced. Our method shows consistent gains across all metrics in both methods.}
\label{tab:application_MOT}
\vspace{-15pt}
\end{table}

\subsection{Application in Multi-Object Tracking (MOT)}
\label{sec:app_mot}
\noindent\textbf{Settings.} Since consistent tracking and reidentification of objects is an important task in MOT, we experiment with two MOT methods (i.e. two-stage detector-based Traktor~\cite{Traktor_2019_ICCV} and YOLOX-based ByteTrack~\cite{bytetrack_ECCV2022}) and incorporate our instance-mask learning in the detector using Eq.~\ref{eq:ifem}. To validate the performance, we evaluate ByteTrack and Tracktor with and without our instance mask learning on the MOT20~\cite{MOT20_arxiv2020} dataset. Complete details of experiments, dataset, and evaluation metrics are outlined in Appendix~\textcolor{red}{D.1}.

\noindent\textbf{Results.}
As summarized in Table~\ref{tab:application_MOT}, our proposed instance mask learning has significantly enhanced the performance of both Tracktor and ByteTrack across nearly all metrics. For instance, the MOTA score improves from \textbf{70.5 to 71.4} in Tracktor and from \textbf{86.4 to 88.1} in ByteTrack. 
These remarkable improvements suggest that exploiting instance mask information temporally not only enhances VOD but also significantly boosts MOT. Moreover, these findings outline the promising potential of our approach in other video understanding tasks~\cite{Video_instance_segmentation_ICCV2019, person_reid_TPAMI2021, video_object_seg_TPAMI2022_survey}. Qualitative analysis is presented in Appendix~\textcolor{red}{D.2}.

\section{Conclusion and Discussion}
\label{sec:conclusion}


This paper introduces a novel paradigm for video object detection through instance mask-based feature aggregation, refining the process to enhance object understanding across video frames. Extensive experiments on multiple benchmarks with different VOD and MOT methods validate our approach's effectiveness and highlight its potential to advance video understanding. Integrating instance mask learning into video understanding tasks opens novel research opportunities, especially when mask data is unavailable. Future work will explore unifying VOD, MOT, and video instance segmentation~\cite{Video_instance_segmentation_ICCV2019} into a cohesive framework.

\section*{Acknowledgements}
This work was in parts supported by the EU Horizon Europe Framework under grant agreements 101135724 (LUMINOUS) and 101092312 (AIRISE).



{\small
\bibliographystyle{ieee_fullname}
\bibliography{egbib}
}


\end{document}


\twocolumn[{
   \renewcommand\twocolumn[1][]{#1}%
    
   \maketitle
   \vspace{-30pt} 
   \begin{center}
    \end{center}

   \vspace{5mm} 
}]








\renewcommand{\thesection}{\Alph{section}}

\renewcommand{\thesubsection}{\thesection.\arabic{subsection}} 

\renewcommand{\thefigure}{\Roman{figure}} 
\renewcommand{\thetable}{\Roman{table}} 


\tableofcontents

\section{Additional Implementation Details}
\label{sec:FAIM_design}

\subsection{Implementation Details of FAIM}
\label{subsec:imp_details}
\noindent\textbf{Training.} To establish a direct comparison with YOLOV~\cite{YOLOV_AAAI2023}, we employ an identical training strategy and adhere to the original codebase\footnote{\url{https://github.com/YuHengsss/YOLOV}} provided by the authors. One-tenth of the frames from the ImageNet VID training set are sampled to mitigate redundancy. The base detectors are trained using an SGD optimizer for 7 epochs with a batch size of 16 on 2 GPUs, following the protocols in~\cite{YOLOV_AAAI2023}. In line with~\cite{yolox_arxiv2021}, we implement the same cosine learning schedule, reserving first epoch for warm-up and omitting data augmentations in the final two epochs.

Upon integrating base detectors into our FAIM framework, we fine-tune them with a batch size of 16 on a single GPU. Emulating the approach in~\cite{YOLOV_AAAI2023}, we apply a warm-up strategy for the first 15K iterations and continue with the cosine learning rate schedule thereafter. It is worth noting that only the newly introduced video object branch, instance feature extraction module, FCN mask head, and the multi-head attentions are fine-tuned, facilitating a direct comparison with~\cite{YOLOV_AAAI2023}. However, thanks to our adaptable mask prediction branch, a stronger baseline can be trained through transforming YOLOX~\cite{yolox_arxiv2021} into a novel instance segmentation model performing both detection and instance segmentation, simaltenously~\cite{Mask_RCNN_ICCV2017, cascade_maskr_rcnn_TPAMI2019,HTC_2019_CVPR}. We anticipate that fine-tuning such models will yield further enhancements. During training in the Feature Prediction Selection Module (FPSM), the Non-Maximum Suppression (NMS) threshold is set to 0.75 for selecting box predictions ($b$) and corresponding features. In the Temporal Instance Classification Aggregation Module (TICAM), the number of frames $m$ is fixed at 16. Images are randomly resized within a range from 352 $\times$ 352 to a maximum of 672 $\times$ 672, with a stride of 32. Note that since we could not reproduce the results of~\cite{YOLOV_AAAI2023} with stronger data augmentations, we omit this experiment for both YOLOV and our FAIM for direct comparison. 

To generate pseudo ground truth instance masks, we utilize ground truth bounding boxes, which are readily available across video object detection~\cite{Imagenet_InternationalJournalofComputerVision2015, Epic_kitchens_ECCV2018} and multi-object tracking datasets~\cite{mot_paper_cvpr_2019}. We explore multiple box-to-segmentation methods~\cite{boxinst_2021_CVPR, lbox2mask_TPAMI_2024, SAM_2023} capable of deriving instance masks from bounding boxes. We opt for the state-of-the-art pre-trained Box2Mask~\cite{lbox2mask_TPAMI_2024} with a ResNet-101 backbone~\cite{Deep_residual_learning_CVPR_2016}, as well as the universal zero-shot image segmentation method SAM~\cite{SAM_2023}, which uses a pre-trained ViT-H image encoder~\cite{VIT_ICLR2021}. For Box2Mask, we conduct straightforward inference on the training set and align instance masks to the ground truth boxes by maximizing IoU overlap. With SAM, we use ground truth boxes as prompts to produce instance masks.

\vspace{5pt}
\noindent\textbf{Inference.}
The inference stage of FAIM is straightforward. Since the generation of pseudo instance masks is unnecessary for the VOD task, we omit both the \colorbox{green!20}{Pseudo Masks Generator} and our mask prediction \colorbox{green!20}{branch} during inference (refer to Fig.~4 in the main paper). Nonetheless, the learned Instance Feature Extraction Module (IFEM) continues to produce high-quality instance mask features.~We propagate these features along with object classification features to enhance the temporal feature aggregation in TICAM for final predictions.~This simple modification ensures high-quality detections and real-time inference speed at the same time.
For testing, the images are uniformly resized to 576 $\times$ 576, and the NMS threshold is adjusted to 0.5 in FPSM to select more high-quality candidates from the converged model. 


\subsection{Instance Mask Learning and its Applications to Other VOD Methods}
\label{subsec:mask_learning}
\noindent\textbf{Overview.} This section first discusses the computation of mask loss during the training phase. Then, it explains the integration of the instance mask learning into other two-stage video object detection methods.

\subsubsection{Instance Mask Loss}
\label{subsubsec:instance_mask_loss}
After filtering, we obtain \( M' \) predicted masks, as explained in Eq. 5 of the main paper. These masks are resized to match the dimensions of the corresponding ground truth bounding boxes in the image through interpolation. Similarly, the ground truth instance masks, initially the same size as the image, are cropped to the area of their respective bounding boxes. For each predicted mask \( m_i \) in \( M' \), we identify its corresponding ground truth target \( g_i \) by computing the mask Intersection over Union (IoU), as described in \cite{Mask_RCNN_ICCV2017}. The loss between the predicted masks \( M' \) and the ground truth masks \( G \) is then computed using a simple Binary Cross-Entropy (BCE) loss, which is formulated as follows:
\begin{equation}
\text{Instance Mask Loss ($L_{mask}$)} = \frac{1}{N'} \sum_{i=1}^{N'} \text{BCE}(m_i, g_i)    \tag{II}
\end{equation}

Where \( N' \) is the number of predicted masks after filtration. This mask loss $L_{mask}$ is integrated with the detection loss of the base detector YOLOX. Thus, the overall loss function for our model combines the detection loss \( L_{det} \) from YOLOX and the instance mask loss \( L_{mask} \), optimized in an end-to-end multi-task fashion. The total loss function is given by:
\begin{equation}
L_{total} = L_{det} + \lambda L_{mask} \tag{III}
\label{eq:total_loss}
\end{equation}
Here, \( \lambda \) is a balancing parameter ($\lambda$ =1 by default) that controls the contribution of the mask loss to the total loss. This multi-task training approach allows the model to leverage synergies between object detection and instance segmentation. Furthermore, the instance mask loss ensures that the loss computation is class-aware and focuses on learning better class-specific instance mask features, enhancing the overall performance of the video object detector.

\subsubsection{Applications to other VOD Methods}
\label{subsubsec:app_vod}
Following the principals outlined in \S~\textcolor{red}{3.1}, we explore the adaptability and efficacy of our instance mask-based feature aggregation in other two-stage VOD methods, specifically SELSA~\cite{Sequence_level_semantics_aggregation_ICCV2019} and TROIA~\cite{temporal_roi_align_AAAI2021}. Both methods utilize proposal-based feature aggregation, producing aggregated RoI features \(X_{agg}\), as described in Eq.~\textcolor{red}{2} of the main paper. We integrate our mask prediction branch (illustrated in Fig.~\textcolor{red}{4} of the paper) into the training phase of these methods. This integration involves pooling instance mask features and forwarding them to our FCN Mask Head for mask prediction. Like in FAIM, we do not use the mask branch during inference. However, incorporating the instance mask loss during training encourages \(X_{agg}\) to emphasize instance mask-level features, leading to enhanced performance.

For implementation, we adhere to the experimental settings specified in the SELSA config\footnote{\url{https://github.com/open-mmlab/mmtracking/blob/master/configs/vid/selsa/selsa_faster_rcnn_r50_dc5_1x_imagenetvid.py}} for SELSA~\cite{Sequence_level_semantics_aggregation_ICCV2019} and the TROIA config\footnote{\url{https://github.com/open-mmlab/mmtracking/blob/master/configs/vid/temporal_roi_align/selsa_troialign_faster_rcnn_r50_dc5_7e_imagenetvid.py}} for TROIA~\cite{temporal_roi_align_AAAI2021}, as provided in MMTracking~\cite{mmtrack2020}. We use a ResNet-50 backbone network for both methods. We employ the same reference frame sampling strategy as the original baselines for evaluation to ensure a direct comparison.

\begin{figure*}[ht]
  \centering
  \includegraphics[width=\linewidth]{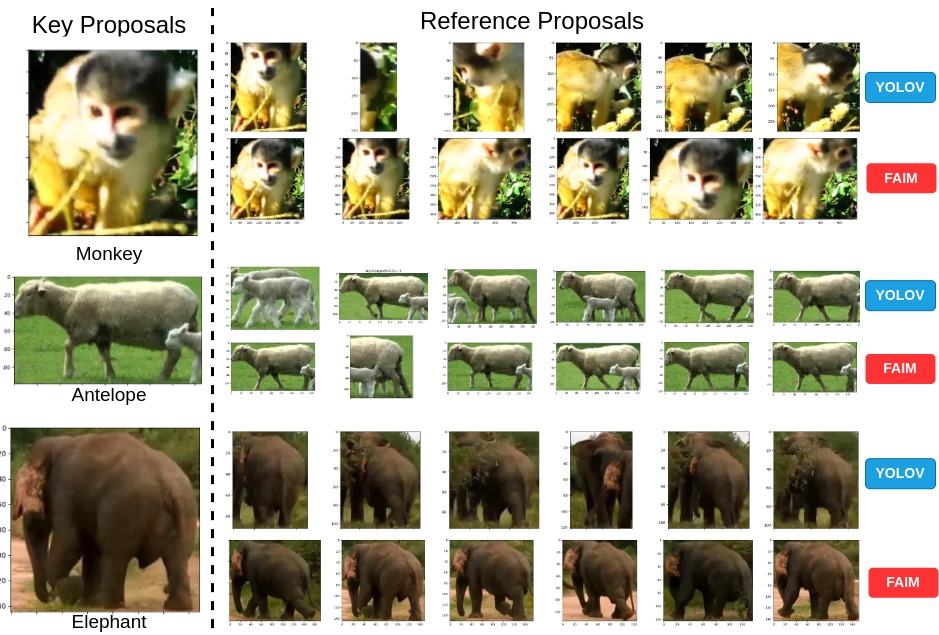}
    \caption{\textbf{Visual comparison of reference proposals from YOLOV~\cite{YOLOV_AAAI2023} and FAIM.} The first row shows the top 6 reference proposals from YOLOV for each key proposal, while the second row displays those from FAIM. YOLOV, using box-level proposals, often selects lower-quality reference proposals, such as background elements in the case of \textbf{\textit{Monkey}} or partially occluded proposals for \textbf{\textit{Elephant}}. On the other hand, FAIM, utilizing instance mask-based learning to reduce background noise, consistently chooses higher-quality reference proposals, enhancing feature aggregation.}
  \label{fig:ref_proposals}
\end{figure*}

\subsection{Details for EPIC-KITCHEN-55}
\label{subsec:imp_details_epic}
EPIC-KITCHENS~\cite{Epic_kitchens_ECCV2018} is a large-scale egocentric dataset that captures daily activities in kitchen environments. Each frame in the dataset contains an average of 1.7 objects and a maximum of 9 objects, presenting a significantly more complex and challenging scenario for video object detection. The task involves 32 different kitchens, encompassing 454,255 object bounding boxes across 290 classes. For training, 272 video sequences captured in 28 kitchens are utilized. The evaluation set comprises 106 sequences collected from the same 28 kitchens (S1) and 54 sequences from 4 additional, unseen kitchens (S2). Videos in the dataset are sparsely annotated at 1-second intervals, making it a complex VOD task.

Following the implementation details in Section~\ref{subsubsec:app_vod}, we incorporate our instance mask-based feature aggregation approach in TROIA~\cite{temporal_roi_align_AAAI2021}. We employ a ResNet-101~\cite{Deep_residual_learning_CVPR_2016} backbone network and adopt identical experimental settings and dataset splits as in~\cite{temporal_roi_align_AAAI2021} for direct comparison. Results are summarized in Table 7 in the main paper.

\definecolor{blue-violet}{rgb}{0.54, 0.17, 0.89}
\begin{figure*}[ht]
  \centering
  \includegraphics[width=.75\linewidth]{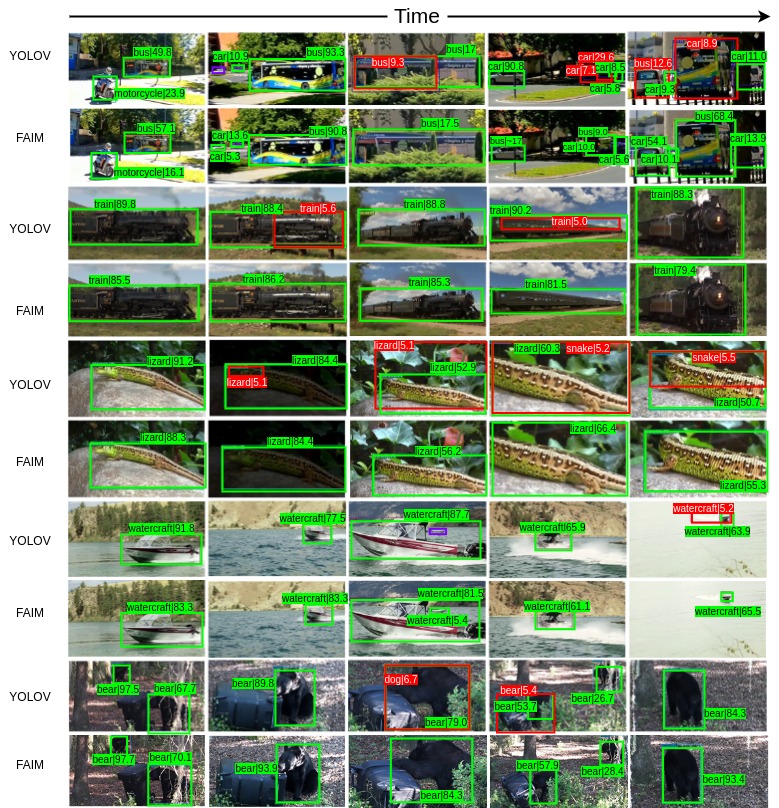}
    \caption{\textbf{Comparing visual performance between FAIM and its counterpart YOLOV~\cite{YOLOV_AAAI2023} on the ImageNet VID validation set.} For each video, the top row presents results from YOLOV, whereas the bottom row denotes results from FAIM. False positives are marked with \textcolor{red}{red boxes} \textcolor{red}{\fbox{}}, and false negatives with \textcolor{blue-violet}{purple boxes} \textcolor{blue-violet}{\fbox{}}. FAIM's instance mask-based feature aggregation effectively reduces intra-class feature variance, leading to more accurate detections. This improvement is particularly noticeable in scenarios with substantial background noise, such as in the cases of \textbf{\textit{Lizard}}, \textbf{\textit{Watercraft}}, and \textbf{\textit{Bear}}, where FAIM outperforms YOLOV in detection accuracy. Best view it on the screen and Zoom in.}
  \label{fig:qual_results}
\end{figure*}

\subsection{Details for OVIS}
\label{subsec:imp_details_ovis}
We also evaluated the capabilities of our FAIM video object detection method on the occluded video instance segmentation (OVIS) dataset~\cite{OVIS_IJCV_2022}. Although OVIS was originally introduced to perform the video instance segmentation task, we obtained its corresponding bounding boxes and classes to train our FAIM, similar to YOLOV~\cite{YOLOV_AAAI2023}. OVIS contains 607 videos for training and 140 for validation, spanning over 25 classes. This dataset contains an average of 4.72 objects per frame, with a large portion suffering from severe occlusions, making it an ideal testbed for our proposed instance mask-based feature aggregation approach.

Since YOLOV~\cite{YOLOV_AAAI2023} is the current state-of-the-art VOD method on this dataset, we draw direct comparisons with it. We employ the base detector YOLOX-X~\cite{yolox_arxiv2021} and evaluate YOLOV-X and FAIM-X on the OVIS dataset, following the experimental settings outlined in~\cite{YOLOV_AAAI2023}. Results are reported in Table 8 in the main paper.

\section{Qualitative Comparisons to Prior Work}
\label{sec:qual_analysis}

\subsection{Inspecting Reference Proposals for Temporal Feature Aggregation}
\label{subsec:proposal_sampling}
We examine the impact of instance mask-based feature aggregation on the selection of reference proposals for feature aggregation. To this end, we compare YOLOV-S and FAIM-S by extracting the top six reference proposals corresponding to the same key proposal. While both methods employ the affinity strategy introduced in~\cite{YOLOV_AAAI2023}, the primary distinction lies in their approach to learning: proposal-based for YOLOV-S and instance mask-based for FAIM-S. As illustrated in Fig.~\ref{fig:ref_proposals}, FAIM's instance mask-based learning significantly enhances the affinity strategy, enabling the selection of higher-quality reference proposals. These proposals are more focused on the target object and exhibit reduced background interference, demonstrating the effectiveness of our approach in refining feature aggregation.

\subsection{Visual Performance Comparison}
\label{subsec:qual_results}
We closely examine the visual detection performance of FAIM and YOLOV~\cite{YOLOV_AAAI2023} across various video sequences on the validation set of ImageNet VID. Fig.~\ref{fig:qual_results} illustrates this comparison, with the detection results from YOLOV presented in the top row and those from FAIM in the bottom row of each video sequence. This visual comparison highlights the effectiveness of FAIM's instance mask-based feature aggregation in enhancing detection accuracy. Notably, FAIM demonstrates a marked improvement in reducing false positives and false negatives, as indicated by the red and purple boxes, respectively. This improvement is especially apparent in challenging scenarios involving significant background noise, such as in the detection of \textit{Lizard}, \textit{Watercraft}, and \textit{Bear}.

\begin{table*}[ht]
\centering
\footnotesize
\begin{tabular}{@{}l|cccc@{}}
\toprule
Model & Mask Source & mAP(\%) & Improvements& Added Training Time \\&&&&(Minutes per Epoch)\\
\midrule
YOLOX-S~\cite{yolox_arxiv2021} & - & 69.5 & -& -\\
YOLOX-L~\cite{yolox_arxiv2021} & - & 76.1 & -& -\\
YOLOX-X~\cite{yolox_arxiv2021} & - & 77.8 & -& -\\
\midrule
FAIM-S  & Box2Mask~\cite{lbox2mask_TPAMI_2024} & 77.9 & +8.4 & +12\\
FAIM-L  & Box2Mask~\cite{lbox2mask_TPAMI_2024} & 84.2 & +8.1& +12\\
FAIM-X  & Box2Mask~\cite{lbox2mask_TPAMI_2024} & 85.5 & +7.7& +12\\
FAIM-S & SAM~\cite{SAM_2023} & 78.2 & +8.7& +26\\
FAIM-L & SAM~\cite{SAM_2023} & 84.3 & +8.2& +26\\
FAIM-X & SAM~\cite{SAM_2023} & 85.6 & +7.8& +26\\
\bottomrule
\end{tabular}
\caption{\textbf{Performance comparison of FAIM models trained with pseudo ground truth instance masks generated by Box2Mask~\cite{lbox2mask_TPAMI_2024} and SAM~\cite{SAM_2023}.} The added training time reflects the additional time per epoch due to different mask sources, calculated on an A100 GPU with a batch size of 4. All other settings remain consistent as described in Sec. 4 of the main paper.}
\label{tab:comparing_sam_box2mask}
\vspace{-10pt}
\end{table*}

\begin{table*}[ht]
\centering

\begin{tabular}{@{}l|cccccc@{}}
Methods & MOTA$\uparrow$ & {IDF1}$\uparrow$ & HOTA$\uparrow$ &{IDS}$\downarrow$ &{FP}$\downarrow$&{FN}$\downarrow$\\
\hline
Tracktor*~\cite{Traktor_2019_ICCV}  & 70.5 & 65.3 & 53.0 & 1442 & 3659 & 176118 \\
Tracktor+Ours  & 
\cellcolor{custom_RoyalBlue!20}\textbf{71.7\textcolor{custom_Turquoise}{\textsubscript{+1.2}}} & \cellcolor{custom_RoyalBlue!20}\textbf{67.5\textcolor{custom_Turquoise}{\textsubscript{+2.2}}} & \cellcolor{custom_RoyalBlue!20}\textbf{53.5\textcolor{custom_Turquoise}{\textsubscript{+0.5}}} & \cellcolor{custom_RoyalBlue!20}\textbf{1307\textcolor{custom_Turquoise}{\textsubscript{-135}}}  & \cellcolor{custom_RoyalBlue!20}\textbf{4003\textcolor{custom_Turquoise}{\textsubscript{+344}}}  & \cellcolor{custom_RoyalBlue!20}\textbf{168595\textcolor{custom_Turquoise}{\textsubscript{-7523}}} \\
\hline
ByteTrack*~\cite{bytetrack_ECCV2022} & 86.4 & 82.7 & 65.5 & 995 & 19176 & 63370 \\
ByteTrack+Ours& 
\cellcolor{custom_RoyalBlue!20}\textbf{86.9\textcolor{custom_Turquoise}{\textsubscript{+0.5}}} & \cellcolor{custom_RoyalBlue!20}\textbf{81.8\textcolor{custom_Turquoise}{\textsubscript{-0.9}}} & \cellcolor{custom_RoyalBlue!20}\textbf{67.0\textcolor{custom_Turquoise}{\textsubscript{+1.5}}} & \cellcolor{custom_RoyalBlue!20}\textbf{974\textcolor{custom_Turquoise}{\textsubscript{-21}}} &
\cellcolor{custom_RoyalBlue!20}\textbf{20938\textcolor{custom_Turquoise}{\textsubscript{+1762}}} &
\cellcolor{custom_RoyalBlue!20}\textbf{58648\textcolor{custom_Turquoise}{\textsubscript{-4722}}}\\
\end{tabular}
\caption{\textbf{ Exploring instance mask-based learning in Multi-Object Tracking with instance masks produced from Box2Mask~\cite{lbox2mask_TPAMI_2024}.} Results with * are reproduced. Our proposed instance mask learning consistently yields marked improvements in both Tracktor and ByteTrack.}
\label{tab:supp_mot_results_box2mask}
\end{table*}

\section{Additional Experiments and Ablations}
\label{sec:more_abl_studies}

\subsection{Results with Pseudo Ground Truth Masks from Box2Mask}
\label{subsec:result_box2mask}
\noindent\textbf{Video Object Detection on ImageNet VID.}
Table~\ref{tab:comparing_sam_box2mask} provides a performance comparison of our FAIM model when trained using pseudo ground truth instance masks generated via two distinct methodologies: Box2Mask and SAM. Significantly, FAIM records notable improvements with both methods, underscoring its adaptability and robustness to the underlying instance mask generation approach. While each method substantially enhances the performance, SAM yields a marginally higher improvement, attributable to its direct utilization of ground truth bounding boxes as prompts for generating instance masks. In contrast, Box2Mask first generates instance masks on the input image, aligned with the ground truth bounding boxes, as elaborated in~\cref{subsec:imp_details}. Nevertheless, the consistent performance gains with both Box2Mask and SAM, highlighted in Table~\ref{tab:comparing_sam_box2mask}, validate that our approach is effectively compatible with any instance mask generation framework, paving the way for further exploration and application in the field.

\noindent\textbf{Multi-Object Tracking. } Table~\ref{tab:supp_mot_results_box2mask} demonstrates the benefits of our instance mask-based learning approach when applied to multi-object tracking, specifically with Tracktor~\cite{Traktor_2019_ICCV} and ByteTrack~\cite{bytetrack_ECCV2022}. We employed Box2Mask~\cite{lbox2mask_TPAMI_2024} to generate pseudo ground truth masks for this evaluation. However, note that Box2Mask is not trained on the MOT20 dataset~\cite{MOT20_arxiv2020}. Nevertheless, we observe noticeable improvements: MOTA increased by +1.2 and +0.5 for Tracktor and ByteTrack, respectively. These enhancements underscore our method's robustness and its independence from the specifics of the mask generation process in MOT. Therefore, any competent box-based instance segmentation tool, even when applied in a zero-shot setting, can complement our framework, affirming the general applicability of our proposed solution.

\subsection{Performance on different motion speeds}
\label{subsec:motion_speed}

Following prior works~\cite{Flow_guided_feature_aggregation_ICCV2017, Sequence_level_semantics_aggregation_ICCV2019, YOLOV_AAAI2023}, we evaluate the detection performance of our FAIM with different motion speeds of objects on the ImageNet VID dataset~\cite{Imagenet_InternationalJournalofComputerVision2015}. The motion speed represents the average Intersection over Union (IoU) scores of objects in consecutive frames. For instance, Slow speed denotes IoU $>$ 0.9, Medium represents 0.9 $\geq$ IoU $\leq$ 0.7, whereas Fast highlights IoU $<$ 0.7. Table~\ref{tab:eval_diff_speed} compares the performance of our FAIM with the image-level feature aggregation method FGFA~\cite{Flow_guided_feature_aggregation_ICCV2017}, proposal-level feature aggregation method SELSA~\cite{Sequence_level_semantics_aggregation_ICCV2019}, the employed base detector YOLOX~\cite{yolox_arxiv2021}, and our direct competitor YOLOV\cite{YOLOV_AAAI2023}. By looking at the results, the effectiveness of our instance mask-based feature aggregation is quite evident, achieving either superior or comparable performance with YOLOV. 

Note that on the smaller model of FAIM-S with the weaker backbone, we achieve a significant gain of +1.1\% in mAP from the previous best results. This boost affirms the benefits of employing instance mask-based feature aggregation, which reduces the intra-class feature variance even in the challenging cases of fast-moving objects.
However, it is crucial to acknowledge that our FAIM-X produces inferior results of -1.4\% in mAP with the stronger detector (YOLOX-X) on the fast motion speed. This outcome is attributed to the fact that our instance mask learning relies on the segmentation qualities of the pseudo mask generator, which occasionally yields under-segmented masks, adversely impacting overall performance.

\begin{table*}[ht]
\centering
\footnotesize
\begin{tabular}{@{}l|cccc@{}}
Methods & Backbone & Slow & Medium & Fast\\
\hline
FGFA~\cite{Flow_guided_feature_aggregation_ICCV2017} & R101 & 87.4 & 79.1 & 61.4  \\
SELSA~\cite{Sequence_level_semantics_aggregation_ICCV2019} & R101 & 88.7 & 83.3 & 71.1  \\
YOLOX-S~\cite{yolox_arxiv2021} & MCSP & 80.1 & 71.4 & 55.3  \\
YOLOX-L~\cite{yolox_arxiv2021} & MCSP & 85.3 & 80.0 & 65.6  \\
YOLOX-X~\cite{yolox_arxiv2021} & MCSP & 87.9 & 80.8 & 68.6  \\
\hline
YOLOV-S~\cite{YOLOV_AAAI2023} & MCSP & 84.6 & 78.6 & 63.7  \\
YOLOV-L~\cite{YOLOV_AAAI2023} & MCSP & 89.3 & 85.8 & 72.6  \\
YOLOV-X~\cite{YOLOV_AAAI2023} & MCSP & 90.6 & 86.8 & 74.8  \\
FAIM-S & MCSP & \cellcolor{custom_RoyalBlue!20}\textbf{84.6}& \cellcolor{custom_RoyalBlue!20}\textbf{79.5} &
\cellcolor{custom_RoyalBlue!20}\textbf{64.8} \\
FAIM-L & MCSP & \cellcolor{custom_RoyalBlue!20}\textbf{89.8}& \cellcolor{custom_RoyalBlue!20}\textbf{86.5} &
\cellcolor{custom_RoyalBlue!20}\textbf{73.0} \\
FAIM-X & MCSP & \cellcolor{custom_RoyalBlue!20}\textbf{91.5}& \cellcolor{custom_RoyalBlue!20}\textbf{86.7} &
\cellcolor{custom_RoyalBlue!20}\textbf{73.4}\\
\hline
\end{tabular}
\caption{\textbf{Evaluating detection performance on different motion speeds.} FAIM either outperforms or is on par with its competitor, YOLOV, across different motion speeds. }
\label{tab:eval_diff_speed}
\end{table*}

\subsection{Upsampling in FCN Mask Head}
\label{subsec:upsampling_ablation}
As detailed in \S~\textcolor{red}{3.2} (in the paper), our FCN Mask head upsamples instance mask features from \(\mathbb{R}^{N \times C \times 32 \times 32}\) to \(\mathbb{R}^{N \times C \times 64 \times 64}\) before applying the final \(1 \times 1\) convolutional layer for prediction. We explore this design choice in Table~\ref{tab:upsampling} and Table~\ref{tab:upsampling_method}, using FAIM-S as the baseline, in line with the ablation studies in Section~\textcolor{red}{4.2}. The results indicate that upsampling via bilinear interpolation yields the most significant improvements. Consequently, this approach is adopted as the default in our FAIM experiments.

\begin{table}[ht]
    \centering

    \begin{tabular}{@{}l|cc@{}}
    Approach $\rightarrow$ & \cellcolor{custom_RoyalBlue!20} Upsampling & No Upsampling  \\
    \hline
    mAP & \cellcolor{custom_RoyalBlue!20} {78.2} & 77.7\\
    \end{tabular}
    \caption{Ablating the effect of upsampling in our FCN Mask Head. Here, upsampling is done with bilinear interpolation.}
    \label{tab:upsampling}
\end{table}

\begin{table}[ht]
    \centering
    \begin{tabular}{@{}l|cc@{}}
    Approach $\rightarrow$ & \cellcolor{custom_RoyalBlue!20} Interpolation & Deconv  \\
    \hline
    mAP & \cellcolor{custom_RoyalBlue!20} {78.2} & 78.0\\
    \end{tabular}
    \caption{Ablation of Different Upsampling Schemes. `Interpolation' refers to bilinear interpolation, while `Deconv' denotes the use of a deconvolutional layer. Bilinear interpolation provides better results. Hence, adopted as default.}
    \label{tab:upsampling_method}
\end{table}

\subsection{FAIM against Attention-based Methods}
\label{subsec:faim_attention_ablation}
To validate the effectiveness of FAIM, we compare it directly with DyHead~\cite{Dynamic_head_CVPR2021}, an attention-based method designed for object detection. DyHead leverages multiple self-attention mechanisms~\cite{transformer_NIPS2017} to enhance scale, spatial, and task awareness. For a fair comparison, we integrate DyHead and our FAIM modules into two strong VOD baselines, SELSA~\cite{Sequence_level_semantics_aggregation_ICCV2019} and TROIA~\cite{temporal_roi_align_AAAI2021}, and evaluate on the ImageNet VID dataset. As shown in Table~\ref{tab:ablation_attention}, FAIM consistently outperforms DyHead, delivering nearly double the improvement. This is primarily due to our proposed TICAM (Temporal Instance and Classification Aggregation Module), which effectively aggregates both object classification queries and instance queries learned from object masks.

\begin{table}[ht]
\centering
\begin{tabular}{@{}l|lll@{}}
Method& {mAP\textsubscript{50}} & {mAP\textsubscript{75}} & {mAP\textsubscript{50:95}} \\
\hline
SELSA~\cite{Sequence_level_semantics_aggregation_ICCV2019}  & 78.4 & 52.5 & 48.6 \\
SELSA+DyHead~\cite{Dynamic_head_CVPR2021} & 78.8\textsubscript{+0.4} & 53.6\textsubscript{+1.1} & 49.0\textsubscript{+0.4} \\
SELSA+FAIM  & \cellcolor{custom_RoyalBlue!20}\textbf{79.5\textcolor{custom_Turquoise}{\textsubscript{+1.1}}} & \cellcolor{custom_RoyalBlue!20}\textbf{54.4\textcolor{custom_Turquoise}{\textsubscript{+1.9}}} & \cellcolor{custom_RoyalBlue!20}\textbf{49.6\textcolor{custom_Turquoise}{\textsubscript{+1.0}}} \\
\hline
TROIA~\cite{temporal_roi_align_AAAI2021}  & 78.9 & 52.8 & 48.8 \\
TROIA+DyHead~\cite{Dynamic_head_CVPR2021} & 79.4\textsubscript{+0.5} & 54.0\textsubscript{+1.2} & 49.3\textsubscript{+0.5} \\
TROIA+FAIM& \cellcolor{custom_RoyalBlue!20}\textbf{80.1\textcolor{custom_Turquoise}{\textsubscript{+1.2}}} & \cellcolor{custom_RoyalBlue!20}\textbf{55.4\textcolor{custom_Turquoise}{\textsubscript{+2.6}}} & \cellcolor{custom_RoyalBlue!20}\textbf{50.0\textcolor{custom_Turquoise}{\textsubscript{+1.2}}} \\
\end{tabular}
\caption{Effectiveness of proposed modules in FAIM against the conventional attention-based method DyHead~\cite{Dynamic_head_CVPR2021} on VOD baselines SELSA~\cite{Sequence_level_semantics_aggregation_ICCV2019} and TROIA~\cite{temporal_roi_align_AAAI2021}. Compared to DyHead,\textbf{ twice the improvement is observed with our proposed FAIM} in both of the VOD baselines.}
\label{tab:ablation_attention}
\end{table}

\subsection{Impact of Instance Mask-Level Aggregation on Higher IoU Thresholds}
\label{subsec:comparison_stric_iou}
We compare the performance of FAIM and YOLOV on the ImageNet VID dataset~\cite{Imagenet_InternationalJournalofComputerVision2015} across different IoU (Intersection over Union) thresholds to substantiate the effectiveness of our proposed finer instance-mask level aggregation. As demonstrated in Table~\ref{tab:higher_iou_thresholds}, FAIM surpasses YOLOV with a significant difference, particularly at higher IoUs such as AP\textsubscript{75} and AP\textsubscript{50:95}. This improvement stems from FAIM's ability to aggregate finer instance mask-level temporal information, leading to enhanced boundary precision and reduced background noise. By capturing finer object details, FAIM provides more accurate bounding box predictions, especially in challenging scenarios requiring higher overlap, thus achieving superior performance at more stringent IoU thresholds.

\begin{table*}[ht]
\centering
\begin{tabular}{@{}l|cccc@{}}
\textbf{Method} & \textbf{AP\textsubscript{50}(\%)$\uparrow$} & \textbf{AP\textsubscript{75}(\%)$\uparrow$} & \textbf{AP\textsubscript{50:95}(\%)$\uparrow$}  & \textbf{Time (ms)$\downarrow$} \\
\hline
YOLOV-S~\cite{YOLOV_AAAI2023} & 77.3 & 60.0 & 54.1 & 11.3 \\
YOLOV-L~\cite{YOLOV_AAAI2023} & 83.6 & 72.0 & 64.2 & 16.3 \\
YOLOV-X~\cite{YOLOV_AAAI2023} & 85.0 & 73.5 & 65.1 & 22.7 \\
FAIM-S & \cellcolor{custom_RoyalBlue!20}\textbf{78.2}\textcolor{custom_Turquoise}{\textsubscript{+0.9}} & \cellcolor{custom_RoyalBlue!20}\textbf{61.7}\textcolor{red}{\textsubscript{+1.7}} & \cellcolor{custom_RoyalBlue!20}\textbf{56.7}\textcolor{red}{\textsubscript{+2.6}} & \cellcolor{custom_RoyalBlue!20}\textbf{11.6} \\
FAIM-L & \cellcolor{custom_RoyalBlue!20}\textbf{84.3}\textcolor{custom_Turquoise}{\textsubscript{+0.7}} & \cellcolor{custom_RoyalBlue!20}\textbf{73.5}\textcolor{red}{\textsubscript{+1.5}} & \cellcolor{custom_RoyalBlue!20}\textbf{66.5}\textcolor{red}{\textsubscript{+2.3}} & \cellcolor{custom_RoyalBlue!20}\textbf{16.5} \\
FAIM-X & \cellcolor{custom_RoyalBlue!20}\textbf{85.6}\textcolor{custom_Turquoise}{\textsubscript{+0.6}} & \cellcolor{custom_RoyalBlue!20}\textbf{74.8}\textcolor{red}{\textsubscript{+1.3}} & \cellcolor{custom_RoyalBlue!20}\textbf{67.9}\textcolor{red}{\textsubscript{+2.8}} & \cellcolor{custom_RoyalBlue!20}\textbf{22.7} \\
\bottomrule
\end{tabular}
\caption{Performance comparison between FAIM and its direct competitor YOLOV on the ImageNet VID dataset across different IoU thresholds. \textbf{FAIM demonstrates even stronger gains at higher IoU thresholds}, due to the proposed finer instance mask-level aggregation.}
\label{tab:higher_iou_thresholds}
\end{table*}


\section{Applications to Multi-Object Tracking}
\label{sec:app_mot}

\subsection{Detailed Settings}
\label{subsec:det_settings}
We investigate the capabilities of our proposed instance mask learning in another important yet challenging video understanding task of multi-object tracking (MOT). For experiments, we replicate two distinct MOT methodologies: the two-stage detector-based Tracktor~\cite{Traktor_2019_ICCV} and the YOLOX-based ByteTrack~\cite{bytetrack_ECCV2022}. This replication adheres to the experimental configurations detailed in Tracktor-R50\footnote{\url{https://github.com/open-mmlab/mmtracking/blob/master/configs/mot/tracktor/tracktor_faster-rcnn_r50_fpn_8e_mot20-private-half.py}\label{fn:tracktor-config}} and ByteTrack-R50\footnote{\url{https://github.com/open-mmlab/mmtracking/blob/master/configs/mot/bytetrack/bytetrack_yolox_x_crowdhuman_mot20-private.py}} in MMTracking framework~\cite{mmtrack2020}.~We conduct experiments on the MOT20~\cite{MOT20_arxiv2020} dataset under the private detection protocol. The MOT20 is a challenging benchmark, comprising pedestrians in crowded scenes with several cases of occlusions. Following the similar approach explained in \S~\ref{subsec:mask_learning}, we generate instance masks for each pedestrian using their corresponding ground truth bounding boxes as prompts for SAM~\cite{SAM_2023}. The primary objective here is not to set new performance benchmarks but to assess the efficacy of our method. To this end, we train our models on a split-half of the MOT20 training set and evaluate them on the remaining half, as detailed in the referenced configuration\footnote{\footref{fn:tracktor-config}}. Following common convention in MOT~\cite{bytetrack_ECCV2022,Traktor_2019_ICCV}, we report the results using the standard CLEAR evaluation metrics~\cite{clear_MOT_metrics_2008}, including MOTA, FP, FN, IDS, IDF1~\cite{IDF1_eccv_2016}, and HOTA~\cite{HOTA_IJCV_2021}.

\begin{figure*}
  \centering
  \includegraphics[width=.7\linewidth]{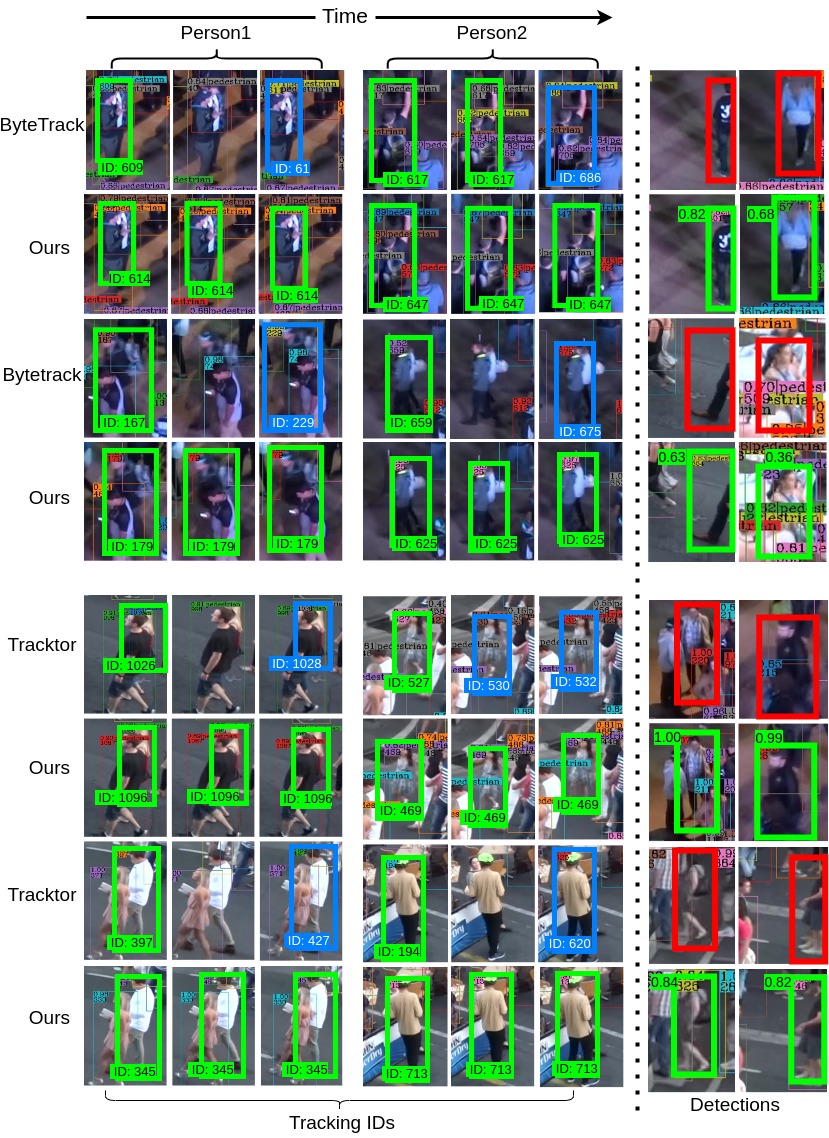}
    \caption{\textbf{Comparing Visual Performance on Tracktor~\cite{Traktor_2019_ICCV} and ByteTrack~\cite{bytetrack_ECCV2022} with and without our proposed instance mask learning on the MOT 20 dataset.} Correct detections and trackings are marked with \textcolor{green}{green bounding boxes} \textcolor{green}{\fbox{}}. \textcolor{cyan}{Blue bounding boxes} \textcolor{cyan}{\fbox{}} indicate tracking errors, while \textcolor{red}{red bounding boxes} \textcolor{red}{\fbox{}} highlight missed detections. Incorporating instance mask learning consistently improves both tracking and detection capabilities in these powerful tracking methods. Best view it on the screen and Zoom in.}
  \label{fig:mot_qual_results}
\end{figure*}
\begin{figure*}[ht]
  \centering
  \includegraphics[width=\linewidth]{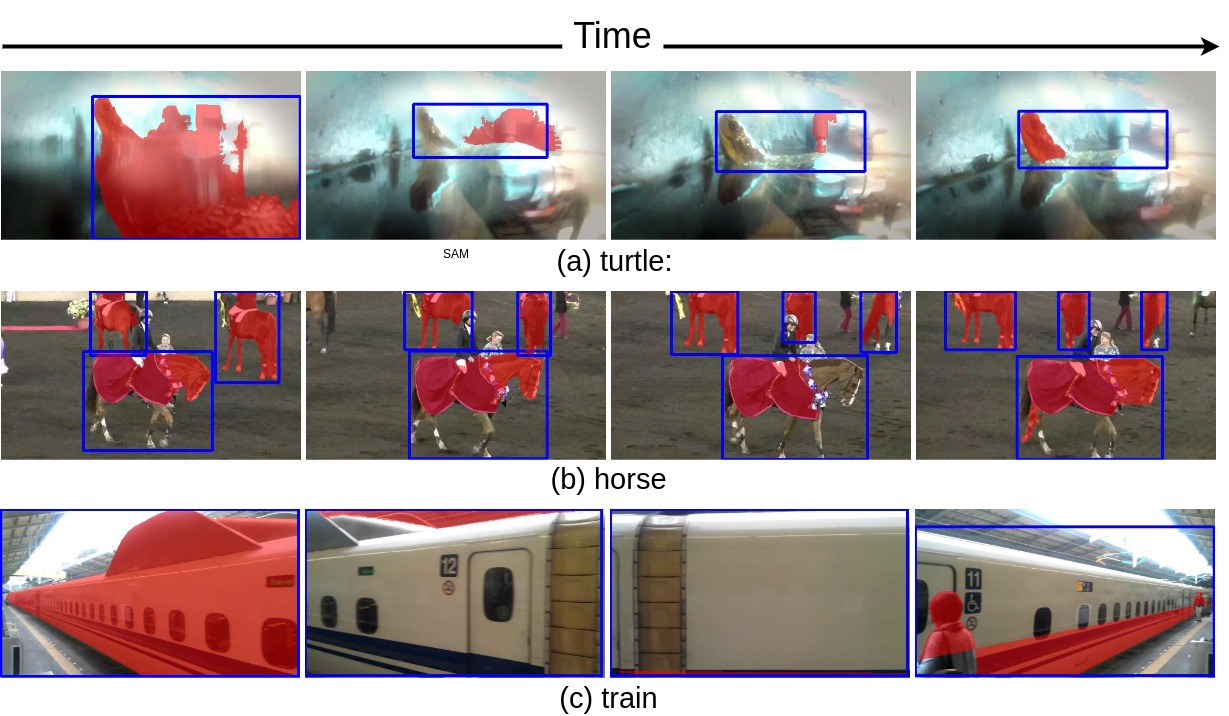}
    \caption{\textbf{Visualizing Incorrectly Generated Instance Masks by SAM~\cite{SAM_2023} and Box2Mask~\cite{lbox2mask_TPAMI_2024} on ImageNet VID Training Set:} \textbf{(a)} SAM sometimes produces under-segmented masks for partially submerged objects. \textbf{(b)} Box2Mask misses or partially captures camouflaged objects. \textbf{(c)} Low-quality masks are produced with SAM when the prompted bounding box encompasses the entire image.}
  \label{fig:sam_limitations}
\end{figure*}

\begin{table*}[ht]
\centering
\begin{tabular}{@{}l|cccccc@{}}
Methods & MOTA$\uparrow$ & {IDF1}$\uparrow$ & HOTA$\uparrow$ &{IDS}$\downarrow$ &{FP}$\downarrow$&{FN}$\downarrow$\\
\hline
Tracktor*~\cite{Traktor_2019_ICCV}  & 70.5 & 65.3 & 53.0 & 1442 & 3659 & 176118 \\
Tracktor+Ours  & 
\cellcolor{custom_RoyalBlue!20}\textbf{71.4\textcolor{custom_Turquoise}{\textsubscript{+0.9}}} & \cellcolor{custom_RoyalBlue!20}\textbf{66.7\textcolor{custom_Turquoise}{\textsubscript{+1.4}}} & \cellcolor{custom_RoyalBlue!20}\textbf{53.1\textcolor{custom_Turquoise}{\textsubscript{+0.1}}} & \cellcolor{custom_RoyalBlue!20}\textbf{1344\textcolor{custom_Turquoise}{\textsubscript{-98}}}  & \cellcolor{custom_RoyalBlue!20}\textbf{3419\textcolor{custom_Turquoise}{\textsubscript{-240}}}  & \cellcolor{custom_RoyalBlue!20}\textbf{171174\textcolor{custom_Turquoise}{\textsubscript{-4944
}}} \\
\hline
ByteTrack*~\cite{bytetrack_ECCV2022} & 86.4 & 82.7 & 65.5 & 995 & 19176 & 63370 \\
ByteTrack+Ours& 
\cellcolor{custom_RoyalBlue!20}\textbf{88.1\textcolor{custom_Turquoise}{\textsubscript{+1.7}}} & \cellcolor{custom_RoyalBlue!20}\textbf{83.7\textcolor{custom_Turquoise}{\textsubscript{+1.0}}} & \cellcolor{custom_RoyalBlue!20}\textbf{68.9\textcolor{custom_Turquoise}{\textsubscript{+3.4}}} & \cellcolor{custom_RoyalBlue!20}\textbf{911\textcolor{custom_Turquoise}{\textsubscript{-84}}}&
\cellcolor{custom_RoyalBlue!20}\textbf{18647\textcolor{custom_Turquoise}{\textsubscript{-529}}}&
\cellcolor{custom_RoyalBlue!20}\textbf{53825\textcolor{custom_Turquoise}{\textsubscript{-9545}}}\\
\end{tabular}
\caption{\textbf{Extended Results of exploring instance mask-based learning in Multi-Object Tracking.} Results with * are reproduced. Our proposed instance mask learning consistently yields significant improvements across all metrics for both Tracktor and ByteTrack.}
\label{tab:supp_mot_results}
\end{table*}

\subsubsection{Incorporating Instance Mask Learning}
\label{subsec:instance_mask_mot}
The integration of instance mask learning into Tracktor and ByteTrack, which utilize distinct object detection algorithms, necessitates tailored approaches. For Tracktor, proposal features are sourced from the RoI head of Faster R-CNN~\cite{Faster_R_CNN_NEURIPS2015}. Subsequently, our mask prediction branch is introduced, initially extracting instance mask features and then generating predictions via our FCN Mask head. Conversely, ByteTrack, based on YOLOX~\cite{yolox_arxiv2021}, does not provide direct access to proposal features. Instead, we harness the classification head features of YOLOX to pool instance mask features. Our FCN Mask head then processes these features for the instance mask prediction. The rationale behind using classification features in ByteTrack is to enhance classification scores through pixel-level learning of instance masks, thereby boosting overall tracking performance. During the training phase for both methods, the instance mask loss is incorporated alongside other losses and trained in a multi-task fashion, as delineated in Eq.~\ref{eq:total_loss}.

\subsection{Performance Analysis}
\label{subsec:qual_analysis}

\subsubsection{Complete Results of Table 7}
\label{subsec:Quantitative_MOT}
Table~\ref{tab:supp_mot_results} provides a detailed summary of the extended results, complementing those in Table~\textcolor{red}{7} of the main paper. As evidenced, remarkably, the integration of our simple instance mask learning into both two-stage and single-stage tracking methods results in substantial improvements across all evaluated metrics.

\subsubsection{Qualitative Analysis}
\label{subsec:Qualitative_mot}
Fig.~\ref{fig:mot_qual_results} offers a detailed qualitative analysis, showcasing the impact of incorporating our instance mask learning into the Tracktor~\cite{Traktor_2019_ICCV} and ByteTrack~\cite{bytetrack_ECCV2022} frameworks. Correctly tracked and detected pedestrians are prominently marked, demonstrating the refined tracking capabilities under complex scenarios, particularly in crowded and occluded environments in the MOT 20 dataset. This qualitative demonstration not only confirms our quantitative findings but also emphasizes the practical benefits and increased robustness of Tracktor and ByteTrack when augmented with our instance mask learning technique.

\section{Limitations and Future Work}
\label{sec:sam_limitations}
While FAIM demonstrates impressive performance in different video understanding tasks, its reliance on zero-shot segmentation methods for the instance mask generation, such as SAM~\cite{SAM_2023} and Box2Mask~\cite{lbox2mask_TPAMI_2024}, introduces potential limitations. These methods produce suboptimal masks in scenarios with complex backgrounds or overlapping objects, as exemplified in Fig.~\ref{fig:sam_limitations}.

Addressing these challenges represents a vital direction for future research. One promising avenue could involve effectively leveraging temporal information within the Pseudo Mask Generator. Furthermore, this integration of instance mask learning into video object tasks, particularly where mask data is not inherently available, paves the way for novel research opportunities and could potentially unify various aspects of video understanding tasks, including video object detection, person re-identification~\cite{person_reid_TPAMI2021}, multi-object tracking~\cite{MOTS_CVPR2019}, video object segmentation~\cite{video_object_seg_TPAMI2022_survey}, video instance segmentation~\cite{Video_instance_segmentation_ICCV2019}, and video panoptic segmentation~\cite{Video_panoptic_segmentation_CVPR2020}.

\section{Ethical Considerations}
\label{sec:ethical}
This work contributes to the advancement of visual recognition and tracking in videos. Although our methodological development does not raise immediate ethical concerns, as with any model, we recommend thorough validation prior to deployment.

{\small
\bibliographystyle{ieee_fullname}
\bibliography{egbib}
}